\documentclass[runningheads]{llncs}
\usepackage{graphicx}

  \makeatletter
  \let\MYcaption\@makecaption
  \makeatother

\usepackage{tikz}
\usepackage{comment}
\usepackage{amsmath,amssymb} 
\usepackage{color}

\usepackage{graphicx}
\usepackage{booktabs}
\usepackage{dsfont}
\usepackage{mathtools}
\usepackage{tabularx}
\usepackage{xcolor,colortbl}
\usepackage{nicefrac}
\usepackage{bm}
\usepackage{makecell}
\usepackage{caption}
\usepackage{comment}
\usepackage{subcaption}

  \makeatletter
  \let\@makecaption\MYcaption
  \makeatother

\usepackage{comment,setspace}
\usepackage[ruled,vlined]{algorithm2e}
\SetArgSty{textnormal}

\colorlet{myGray}{black!5}
\newcolumntype{a}{>{\columncolor{myGray}}c}

\makeatletter
\@namedef{ver@everyshi.sty}{}
\makeatother
\usepackage{tikz,pgfplots}
\pgfplotsset{compat=newest} 
\pgfkeys{/pgf/number format/.cd,1000 sep={}}
\usepgfplotslibrary{groupplots}
\usetikzlibrary{decorations.pathreplacing,shapes.misc}

\newlength\figureheight
\newlength\figurewidth

\pgfplotsset{every axis/.append style={
  grid style={line width=0.6pt,dotted,gray}}}
  
\pgfplotsset{every axis/.append style={
		legend style={inner xsep=1pt, inner ysep=0.5pt, nodes={inner sep=1pt, text depth=0.1em},draw=none,fill=none}
}}

\usetikzlibrary{quotes,shapes.arrows,shapes.geometric,arrows.meta,decorations.markings,decorations.text}

\definecolor{mycolor0}{rgb}{0.2667,0.4471,0.7098}
\definecolor{mycolor1}{rgb}{0.1647,0.6706,0.3804}
\definecolor{mycolor2}{rgb}{0.8275,0.2627,0.3059}
\definecolor{mycolor3}{rgb}{0.5216,0.4392,0.7176}
\definecolor{mycolor4}{rgb}{0.8118,0.7255,0.4118}
\definecolor{mycolor5}{rgb}{0.2745,0.7176,0.8157}

\definecolor{mylcolor0}{rgb}{0.6902,0.7686,0.8863}
\definecolor{mylcolor1}{rgb}{0.5451,0.8902,0.6941}
\definecolor{mylcolor2}{rgb}{0.9412,0.7490,0.7647}
\definecolor{mylcolor3}{rgb}{0.8627,0.8392,0.9176}
\definecolor{mylcolor4}{rgb}{0.9569,0.9373,0.8667}
\definecolor{mylcolor5}{rgb}{0.7529,0.9020,0.9373}
\definecolor{mylcolor6}{rgb}{0.8750,0.8750,0.8750}

\usepackage{hyperref}

\usepackage[capitalize,nameinlink]{cleveref}
\crefname{section}{Sec.}{Secs.}
\crefname{algorithm}{Alg.}{Algs.}

\newcommand{\mbf}[1]{\mathbf{#1}}

\newcommand\inv[1]{#1\raisebox{1.15ex}{$\scriptscriptstyle-\!1$}}
\newcommand\tanspos[1]{#1^\top}
\newcommand{\dd}{\,\mathrm{d}} 

\newcommand{\va}{\mbf{a}}

\newcommand{\vp}{\mbf{p}}

\newcommand{\vx}{\mbf{x}}
\newcommand{\vy}{\mbf{y}}
\newcommand{\vz}{\mbf{z}}

\newcommand{\MH}{\mbf{H}}

\newcommand{\MU}{\mbf{U}}
\newcommand{\MV}{\mbf{V}}

\newcommand{\MX}{\mbf{X}}
\newcommand{\MY}{\mbf{Y}}
\newcommand{\MZ}{\mbf{Z}}

\newcommand{\X}{\mathcal{X}}
\newcommand{\Y}{\mathcal{Y}}
\newcommand{\data}{\mathcal{D}}
\newcommand{\source}{\mathtt{S}}
\newcommand{\target}{\mathtt{T}}
\newcommand{\ttarget}{\ensuremath{^{[\target]}}}
\newcommand{\tsource}{\ensuremath{^{[\source]}}}
\newcommand{\labels}{\mathrm{L}}
\newcommand{\MAP}{\mathrm{MAP}}
\renewcommand{\mid}{\,|\,}
\def\Plus{{\texttt{+}}}

\newcommand{\etal}{\textit{et al.}\xspace}
\newcommand{\eg}{\textit{e.g.}\xspace}
\newcommand{\ie}{\textit{i.e.}\xspace}

\newcommand{\ths}{\textsuperscript{th}\;}

\newcommand{\ours}{U-SFAN\xspace}
\DeclareMathOperator{\E}{\mathbb{E}}

\usepackage{fontawesome5}

\def\tableoptions{\def\arraystretch{.85}}

\renewcommand{\paragraph}[1]{\smallskip\noindent{\bf #1}~~}

\usepackage[accsupp]{axessibility}  

\begin{document}
\pagestyle{headings}
\mainmatter
\def\ECCVSubNumber{6976}  

\title{Uncertainty-guided Source-free \\Domain Adaptation} 

\titlerunning{Uncertainty-guided Source-free Domain Adaptation}
%
\author{Subhankar Roy\inst{1,2} \and
Martin Trapp\inst{3} \and
Andrea Pilzer\inst{4} \and
Juho Kannala\inst{3} \and \\
Nicu Sebe\inst{1} \and
Elisa Ricci\inst{1,2} \and
Arno Solin\inst{3}
}
\authorrunning{S. Roy et al.}
%
\institute{University of Trento, Trento, Italy \and
Fondazione Bruno Kessler, Trento, Italy \and 
Aalto University, Espoo, Finland\and
NVIDIA\\
\email{subhankar.roy@unitn.it}}
\maketitle

\begin{abstract}
Source-free domain adaptation (SFDA) aims to adapt a classifier to an unlabelled target data set by only using a pre-trained source model. However, the absence of the source data and the domain shift makes the predictions on the target data unreliable. We propose quantifying the uncertainty in the source model predictions and utilizing it to guide the target adaptation. For this, we construct a probabilistic source model by incorporating priors on the network parameters inducing a distribution over the model predictions. Uncertainties are estimated by employing a Laplace approximation and incorporated to identify target data points that do not lie in the source manifold and to down-weight them when maximizing the mutual information on the target data. Unlike recent works, our probabilistic treatment is computationally lightweight, decouples source training and target adaptation, and requires no specialized source training or changes of the model architecture. We show the advantages of uncertainty-guided SFDA over traditional SFDA in the closed-set and open-set settings and provide empirical evidence that our approach is more robust to strong domain shifts even without tuning.

\end{abstract}


\section{Introduction}
\label{sec:intro}

Deep neural networks have proven to be very successful in a myriad of computer vision tasks such as categorization, detection, and retrieval. However, much of the success has come at the price of excessive human effort put into the manual data-labelling process. Since collecting annotated data can be prohibitive and impossible at times, domain adaptation (DA, see~\cite{csurka2017comprehensive} for an overview) methods have gained increasing attention. They enable training on unlabelled target data by conjointly leveraging a previously labelled yet related source data set while mitigating \emph{domain-shift}~\cite{torralba2011unbiased} between the two. Such methods predominantly comprise of minimizing statistical moments between distributions~\cite{tzeng2014deep,long2015learning,sun2016deep,roy2019unsupervised}, using adversarial objectives to maximize domain confusion~\cite{ganin2016domain,Hoffman:Adda:CVPR17}, or reconstructing data with generative methods~\cite{hoffman2018cycada}.

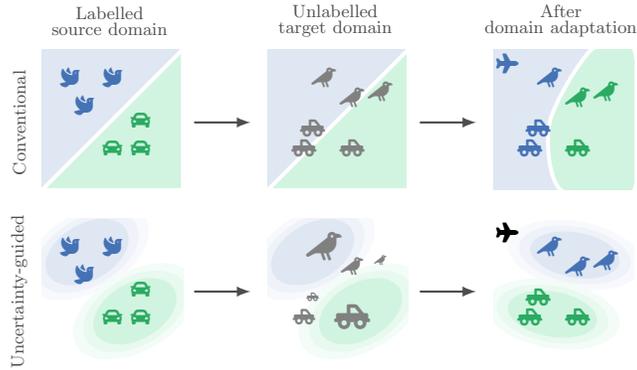
\begin{figure}[!t]
\centering
\tikzset{
    show curve controls/.style={
        decoration={
            show path construction,
            curveto code={
                \draw [blue, dashed]
                    (\tikzinputsegmentfirst) -- (\tikzinputsegmentsupporta)
                    node [at end, cross out, draw, solid, red, inner sep=2pt]{};
                \draw [blue, dashed]
                    (\tikzinputsegmentsupportb) -- (\tikzinputsegmentlast)
                    node [at start, cross out, draw, solid, red, inner sep=2pt]{};
            }
        }, decorate
    }
}
    \begin{tikzpicture}[inner sep=0,outer sep=0,scale=.75,transform shape]
    
    \newcommand{\birds}[2]{%
        \node at (#1-0.25,#2+0.) {\textcolor{mycolor0}{\faDove}};
        \node at (#1+0.25,#2+0.5) {\textcolor{mycolor0}{\faDove}};
        \node at (#1-0.5,#2+0.5) {\textcolor{mycolor0}{\faDove}};
    };
    
    \newcommand{\cars}[2]{%
        \node at (#1+0.25,#2+-0.75) {\textcolor{mycolor1}{\faCar}};
        \node at (#1+0.75,#2+-0.25) {\textcolor{mycolor1}{\faCar}};
        \node at (#1+0.75,#2+-0.75) {\textcolor{mycolor1}{\faCar}};
    };
    
    \newcommand{\sourcemodel}[2]{%
        \fill[mylcolor0!40!white] (#1-1,#2-1.5) rectangle (#1+1.5,#2+1);
        \draw[fill=mylcolor1!40!white, draw=white, line width=1.5] (#1+1.5,#2+1) -- (#1+1.5,#2-1.5) -- (#1-1,#2-1.5) -- cycle;
    };
    
    \newcommand{\bayessourcemodel}[2]{%
        \begin{scope}
            \clip (#1-1,#2-1.5) rectangle (#1+1.5,#2+1);
            
            \newcommand{\ellipsex}{#1-0.2}
            \newcommand{\ellipsey}{#2+0.4}
            \fill[mylcolor0!10!white,rotate around={30:(\ellipsex,\ellipsey)}] (\ellipsex,\ellipsey) ellipse (1.2 and 0.7);
            \fill[mylcolor0!20!white,rotate around={30:(\ellipsex,\ellipsey)}] (\ellipsex,\ellipsey) ellipse (1 and 0.6);
            \fill[mylcolor0!40!white,rotate around={30:(\ellipsex,\ellipsey)}] (\ellipsex,\ellipsey) ellipse (0.8 and 0.5);
            
            \renewcommand{\ellipsex}{#1+0.65}
            \renewcommand{\ellipsey}{#2-0.65}
            \fill[mylcolor1!10!white,rotate around={30:(\ellipsex,\ellipsey)}] (\ellipsex,\ellipsey) ellipse (1.2 and 0.7);
            \fill[mylcolor1!20!white,rotate around={30:(\ellipsex,\ellipsey)}] (\ellipsex,\ellipsey) ellipse (1 and 0.6);
            \fill[mylcolor1!40!white,rotate around={30:(\ellipsex,\ellipsey)}] (\ellipsex,\ellipsey) ellipse (0.8 and 0.5);
        \end{scope}
    };
    
    \newcommand{\sfda}[2]{%
    \begin{scope}
        \clip (#1-1,#2-1.5) rectangle (#1+1.5,#2+1);
        \fill[mylcolor0!40!white] (#1-1,#2-1.5) rectangle (#1+1.5,#2+1);
        \draw[fill=mylcolor1!40!white, draw=white, line width=1.5] plot [smooth cycle] coordinates {(#1+0.75,#2+1) (#1+1.5,#2+1) (#1+1.5,#2-1.5) (#1+0.2,#2-1.5) (#1,#2-0.3)};
    \end{scope}
    };
    
    \newcommand{\usfda}[2]{%
    \begin{scope}
            \clip (#1-1,#2-1.5) rectangle (#1+1.5,#2+1);
            \newcommand{\ellipsex}{#1+0.45}
            \newcommand{\ellipsey}{#2+0.35}
            \fill[mylcolor0!10!white,rotate around={-10:(\ellipsex,\ellipsey)}] (\ellipsex,\ellipsey) ellipse (1.2 and 0.6);
            \fill[mylcolor0!20!white,rotate around={-10:(\ellipsex,\ellipsey)}] (\ellipsex,\ellipsey) ellipse (1 and 0.5);
            \fill[mylcolor0!40!white,rotate around={-10:(\ellipsex,\ellipsey)}] (\ellipsex,\ellipsey) ellipse (0.8 and 0.4);
            
            \renewcommand{\ellipsex}{#1+0.05}
            \renewcommand{\ellipsey}{#2-0.7}
            \fill[mylcolor1!10!white,rotate around={-10:(\ellipsex,\ellipsey)}] (\ellipsex,\ellipsey) ellipse (1.2 and 0.6);
            \fill[mylcolor1!20!white,rotate around={-10:(\ellipsex,\ellipsey)}] (\ellipsex,\ellipsey) ellipse (1 and 0.5);
            \fill[mylcolor1!40!white,rotate around={-10:(\ellipsex,\ellipsey)}] (\ellipsex,\ellipsey) ellipse (0.8 and 0.4);
    \end{scope}
    };
    
    \newcommand{\targetbirds}[2]{%
        \node at (#1+0.5,#2+0.15) {\textcolor{gray}{\faCrow}};
        \node at (#1+1,#2+0.25) {\textcolor{gray}{\faCrow}};
        \node at (#1+0,#2+0.5) {\textcolor{gray}{\faCrow}};
    };
    
    \newcommand{\targetbirdsScaled}[2]{%
        \node at (#1+0.5,#2+0.15) {\textcolor{gray}{\small\faCrow}};
        \node at (#1+1,#2+0.25) {\textcolor{gray}{\tiny\faCrow}};
        \node at (#1+0,#2+0.5) {\textcolor{gray}{\Large\faCrow}};
    };
    
    \newcommand{\targetbirdsCorrect}[2]{%
        \node at (#1+0.5,#2+0.15) {\textcolor{mycolor0}{\faCrow}};
        \node at (#1+1,#2+0.25) {\textcolor{mycolor0}{\faCrow}};
        \node at (#1+0,#2+0.5) {\textcolor{mycolor0}{\faCrow}};
    };
    
    \newcommand{\targetbirdsSfda}[2]{%
        \node at (#1+0.5,#2+0.15) {\textcolor{mycolor1}{\faCrow}};
        \node at (#1+1,#2+0.25) {\textcolor{mycolor1}{\faCrow}};
        \node at (#1+0,#2+0.5) {\textcolor{mycolor0}{\faCrow}};
    };
    
    \newcommand{\targetcars}[2]{%
        \node at (#1-0.35,#2-0.75) {\textcolor{gray}{\faTruckPickup}};
        \node at (#1-0.2,#2-0.4) {\textcolor{gray}{\faTruckPickup}};
        \node at (#1+0.5,#2-0.75) {\textcolor{gray}{\faTruckPickup}};
    };
    
    \newcommand{\targetcarsCorrect}[2]{%
        \node at (#1-0.35,#2-0.75) {\textcolor{mycolor1}{\faTruckPickup}};
        \node at (#1-0.2,#2-0.4) {\textcolor{mycolor1}{\faTruckPickup}};
        \node at (#1+0.5,#2-0.75) {\textcolor{mycolor1}{\faTruckPickup}};
    };
    
    \newcommand{\targetcarsScaled}[2]{%
        \node at (#1-0.35,#2-0.75) {\textcolor{gray}{\small\faTruckPickup}};
        \node at (#1-0.2,#2-0.4) {\textcolor{gray}{\tiny\faTruckPickup}};
        \node at (#1+0.5,#2-0.75) {\textcolor{gray}{\Large\faTruckPickup}};
    };
    
    \newcommand{\targetcarsSfda}[2]{%
        \node at (#1-0.35,#2-0.75) {\textcolor{mycolor0}{\faTruckPickup}};
        \node at (#1-0.2,#2-0.4) {\textcolor{mycolor0}{\faTruckPickup}};
        \node at (#1+0.5,#2-0.75) {\textcolor{mycolor1}{\faTruckPickup}};
    };
    
    \newcommand{\targetood}[2]{%
        \node at (#1-0.75,#2+0.75) {\textcolor{black}{\faPlane}};
    };

    \newcommand{\targetoodSfda}[2]{%
        \node at (#1-0.75,#2+0.75) {\textcolor{mycolor0}{\faPlane}};
    };
    
    \sourcemodel{0}{0}
    \birds{0}{0}
    \cars{0}{0}
    
    \sourcemodel{4}{0}
    \targetbirds{4}{0}
    \targetcars{4}{0}
    
    \sfda{8}{0}
    \targetbirdsSfda{8}{0}
    \targetcarsSfda{8}{0}
    \targetoodSfda{8}{0}
    
    \bayessourcemodel{0}{-3}
    \birds{0}{-3}
    \cars{0}{-3}
    
    \bayessourcemodel{4}{-3}
    \targetbirdsScaled{4}{-3}
    \targetcarsScaled{4}{-3}
    
    \usfda{8}{-3}
    \targetbirdsCorrect{8}{-3}
    \targetcarsCorrect{8}{-3}
    \targetood{8}{-3}
    
    \tikzstyle{label} = [text width=3cm,align=center,xshift=2mm,color=black!70]
    \node[label] at (0,1.5) {\footnotesize Labelled\\[-4pt] source domain};
    \node[label] at (4,1.5) {\footnotesize Unlabelled\\[-4pt] target domain};
    \node[label] at (8,1.5) {\footnotesize After\\[-4pt] domain adaptation};    

    \tikzstyle{label} = [text width=3cm,align=center,yshift=-3mm,rotate=90,color=black!70]
    \node[label] at (-1.4,0) {\footnotesize Conventional};
    \node[label] at (-1.4,-3) {\footnotesize Uncertainty-guided};
 
    \tikzstyle{larrow} = [-latex,thick,black!70]
    \draw[larrow] (1.7,-.3) -- (2.7,-.3);
    \draw[larrow] (1.7,-3.3) -- (2.7,-3.3);
    \draw[larrow] (5.7,-.3) -- (6.7,-.3);
    \draw[larrow] (5.7,-3.3) -- (6.7,-3.3);
        
    \end{tikzpicture}
    \caption{Illustrative sketch of source-free domain adaptation (SFDA) on a labelled source domain (\textcolor{mycolor0}{\small\faDove}, \textcolor{mycolor1}{\small\faCar)} and an \emph{unlabelled} target domain (\textcolor{gray}{\small\faCrow},~\textcolor{gray}{\small\faTruckPickup}) potentially containing additional classes (\textcolor{black}{\small\faPlane}). The {\bf top-row} shows conventional methods which ignore model uncertainties; the {\bf bottom-row} shows our method which incorporates uncertainties about the predictive model, enabling uncertainty-guided SFDA that is more robust to distribution shifts}
    \label{fig:teaser}
\end{figure}

Albeit successful, the preceding methods mandate access to the source data set during the target adaptation phase as they require an estimate of the source distribution for the alignment. With the emergence of regulations on data privacy and bottleneck in data transmission for large data sets, access to the source data can not always be guaranteed. Thus, paving the way to a relatively new and more realistic DA setting, called \emph{source-free} DA (SFDA,~\cite{csurka2017comprehensive}), where the task is to adapt to the target data set when the only source of supervision is a source-trained model. SFDA facilitates maintaining data anonymity in privacy-sensitive applications (\eg, surveillance or medical applications) and at the same time reduces data transmission and storage overhead. Towards this goal, recently, several SFDA methods have been proposed that utilize the hypotheses learned from the source data~\cite{liang2020we,tian2021vdm,kundu2020universal}. Notably, SHOT~\cite{liang2020we} -- an information maximization (IM)~\cite{gomes2010discriminative} based SFDA method -- has demonstrated to work reasonably well on DA benchmarks, sometimes outperforming traditional DA methods. While promising, these conventional SFDA techniques do not account for the uncertainty in the predictions of the source model on the target data. As a by-product, solely maximizing mutual information~\cite{gomes2010discriminative} on the target data can lead to erroneous decision surfaces (see~\cref{fig:teaser} top). 

This work argues that quantification of the uncertainty in predictions is essential in SFDA. Depending on the inductive biases of the model, the source model may predict incorrect target pseudo-labels with high confidence, \eg, due to the extrapolation property in ReLU networks~\cite{hein2019relu} (see~\cref{fig:ood} left). In the literature, uncertainty-guided methods have been proposed in the context of traditional UDA and SFDA settings, employing Monte Carlo (MC) dropout to estimate the uncertainties in the model predictions~\cite{ringwald2020unsupervised,zheng2021rectifying}. However, MC dropout requires specialized training and specialized model architecture, suffers from manual hyperparameter tuning~\cite{gal2017concrete}, and is known to provide a poor approximation even for simple (\eg, linear) models~\cite{osband2016risk,osband2018randomized,foong2020expressiveness}.

In this work, we propose to construct a probabilistic source model by incorporating priors on the network parameters, inducing a distribution over the model predictions, on the last layer of the source model. This enables us to perform an efficient local approximation to the posterior using a \textit{Laplace approximation} (LA,~\cite{tierney1986accurate,mackay2003information}), see \cref{fig:laplace}.
This principled Bayesian treatment leads to more robust predictions, especially when the target data set contains out-of-distribution (OOD) classes (see \cref{fig:teaser} bottom) or in case of strong domain shifts.  
Once the uncertainty in predictions is estimated, we selectively guide the target model to maximize the mutual information~\cite{gomes2010discriminative} in the target predictions. This alleviates the alignment of the target features with the wrong source hypothesis, resulting in a domain adaptation scheme that is robust to mild and strong domain shifts without tuning. We call our proposed method \textbf{U}ncertainty-guided \textbf{S}ource-\textbf{F}ree \textbf{A}daptatio\textbf{N} (U-SFAN). Our approach requires no specialized source training or specialized architecture, opposed to exiting works (\eg, \cite{lao2020hypothesis,zheng2021rectifying}), introduces little computational overhead, and decouples source training and target adaptation.

We summarize our contributions as follows. {\em (i)}~We emphasize the need to quantify uncertainty in the predictions for SFDA and propose to account for uncertainties by placing priors on the parameters of the source model. Our approach is computationally efficient by employing a last-layer Laplace approximation and greatly decouples the training of the source and target.
{\em (ii)}~We demonstrate that our proposed \ours successfully guides the target adaptation without specialized loss functions or a specialised architecture.
{\em (iii)}~We empirical show the advantage of our method over SHOT \cite{liang2020we} in the closed-set and the open-set setting for several benchmarks tasks and provide evidence for the improved robustness against mild and strong domain shifts.


\section{Related Work}
\label{sec:related}

\begin{figure*}[t!]
    \centering
    \begin{subfigure}[t]{0.5\textwidth}
        \centering\scriptsize
        \setlength{\figurewidth}{.9\columnwidth}
        \setlength{\figureheight}{.5\figurewidth}
        \pgfplotsset{axis on top,scale only axis,ytick style={draw=none},xtick style={draw=none},yticklabels={},axis lines = middle,y label style={at={(axis description cs:0,.5)},rotate=90,anchor=south}}
        \input{./figs/laplace.tex}
        \caption{Laplace approximation}
        \label{fig:laplace}
    \end{subfigure}%
    ~ 
    \begin{subfigure}[t]{0.5\textwidth}
        \centering
        \includegraphics[width=\columnwidth]{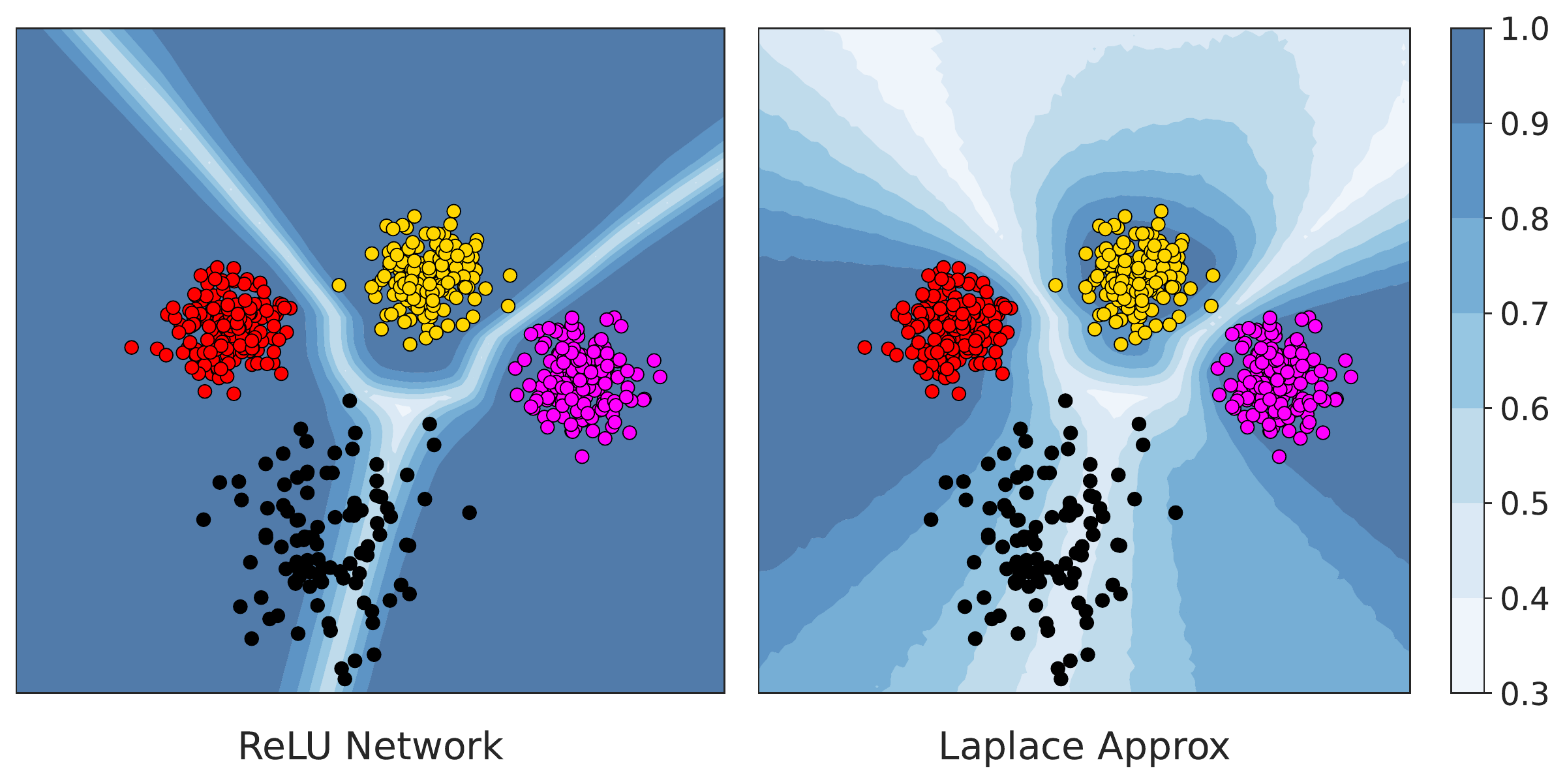}
        \caption{Out-of-distribution detection}
        \label{fig:ood}
    \end{subfigure}
    \newcommand{\foo}{\protect\tikz[baseline=-.3ex]{\protect\node[circle,fill=black,inner sep=.8pt]{};\protect\node[circle,fill=black,inner sep=.8pt,yshift=3pt,xshift=2pt]{};\protect\node[circle,fill=black,inner sep=.8pt,yshift=-0.5pt,xshift=4pt]{};}}
    \caption{(a)~The Laplace approximation is mode-seeking and adapts to the local curvature around the mode $\theta_\mathrm{MAP}$. It does not necessarily capture the (intractable) full posterior, but gives a proxy for it, is principled, and efficient to evaluate. (b)~Example of predictive uncertainty (un~\includegraphics[width=16pt,height=6pt]{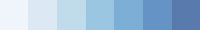}~certain) captured by a ReLU network vs.\ a Laplace approximation that assigns higher uncertainty to inputs (\foo) of an unseen class}
    \label{fig:laplace-ood}
\end{figure*}

\textbf{Closed-set Domain Adaptation}, often abbreviated as UDA, refers to the family of DA methods that aim to learn a classifier for an unlabelled target data set while simultaneously using the labelled source data set, which differ in their underlying data distributions. In the literature~\cite{wang2018deep} mainly three categories of UDA methods can be found. First, discrepancy-based UDA methods aim to diminish the domain-shift between the two domains with maximum mean discrepancy (MMD,~\cite{tzeng2014deep,long2015learning,long2017deep}), or with correlation alignment~\cite{sun2016deep,morerio2017minimal,roy2019unsupervised}. The second category of UDA methods exploits the adversarial objective~\cite{goodfellow2014generative} to promote domain confusion between the two data distributions by using domain discriminator~\cite{ganin2016domain,Hoffman:Adda:CVPR17,long2018conditional}. Finally, the third category comprises reconstruction-based UDA methods~\cite{bousmalis2016domain,ghifary2016deep,hoffman2018cycada} that casts data reconstruction as an auxiliary objective in order to ensure invariance in the feature space. However, these methods can only work in the presence of the source data set during the adaptation stage, which might be limited in practice due to data privacy or storage concerns.

\textbf{Open-set Domain Adaptation} (OSDA), originally proposed in~\cite{panareda2017open}, refers to the DA setting where both the domains have some shared and private classes, with explicit knowledge about the shared classes. However, such a setting was deemed impractical, and later Saito \etal~\cite{saito2018open} proposed the open-set setting where the source labels are a subset of the target labels. Thereon, several OSDA methods have been proposed which use image-to-image translations~\cite{zhang2019improving}, progressive filtering~\cite{liu2019separate}, ensemble of multiple classifiers~\cite{fu2020learning} and one-vs-all classifiers~\cite{saito2021ovanet} to detect OOD samples. Similar to the UDA, the OSDA methods also require support from the source data to detect target private classes, which make them unsuitable for source-free DA.

\textbf{Source-free Domain Adaptation} (SFDA) aims to adapt a model to the unlabelled target domain when only the source model is available and the source data set is absent during target adaptation. Existing SFDA methods use pseudo-label refinement~\cite{liang2020we,ahmed2021adaptive,chen2021self}, latent source feature generation using variational inference~\cite{yeh2021sofa}, or disparity among an ensemble of classifiers~\cite{lao2020hypothesis}. Certain SFDA methods resort to {\it ad hoc} source training protocols to enable the source model to be adapted on the target data. For instance,~\cite{lao2020hypothesis} requires an ensemble of classifiers to be trained during source training so that the disparity among them could be utilized for target adaptation. Similarly, USFDA~\cite{kundu2020universal} requires artificially generated negative samples in the source training stage for the model to detect OOD samples. Such coupled source and target training procedures make these SFDA methods less viable for practical applications. On the other hand, our proposed \ours does not require specialized source training except a computationally lightweight approximate inference, which can be done with a single pass of the source data during the source training. Moreover, unlike~\cite{lao2020hypothesis,ahmed2021adaptive}, our \ours works well on both closed-set and open-set SFDA without {\it ad~hoc} modifications.

\textbf{Uncertainty Quantification} in the form of Bayesian deep learning (\eg, \cite{neal2012bayesian,kendall2017what}) is concerned with formalizing prior knowledge and representing uncertainty in model predictions, especially under domain-shift or out-of-distribution samples. Even though the Bayesian methodology gives an explicit way of formalizing uncertainty, computation is often intractable. Thus, approximate inference methods such as Monte Carlo (MC) dropout~\cite{gal2016dropout}, deep ensembles~\cite{lakshminarayanan2016simple,Wilson:ensembles}, other stochastic methods (\eg,~\cite{maddox19_SWAG}), variational methods \cite{BleiKM16}, or the Laplace approximation~\cite{ritter2018scalable} are typically employed in practice.
Prior works in semantic segmentation  \cite{zheng2021rectifying} and  UDA \cite{ringwald2020unsupervised,lao2020hypothesis,wen2019bayesian,han2019unsupervised,kurmi2019attending} applied MC dropout or deep ensembles, respectively, for uncertainty quantification if DA.
However, none of those above approaches can be considered practical for the more challenging source-free DA scenario as MC dropout, ensembles, and other stochastic methods do not lend themselves well to the source-free case. In particular, they either require retraining several models on the source, changing the model architecture or requiring a tailored learning procedure on the source data. Thus we take a Laplace approach which allows re-using the source model by linearizing around a point-estimate (see \cref{fig:laplace-ood}), which is {\it post~hoc}, yet grounded in classical statistics~\cite{gelman2013bayesian}.

\section{Methods}
\label{sec:method}

\paragraph{Problem Definition and Notation} We are given a labelled source data set, having $n\tsource$ instances, $\data\tsource = \{(\vx\tsource_i, \vy\tsource_i)\}^{n\tsource}_{i=1}$, where $\vx\tsource \in \X\tsource$ are $D$-dimensional inputs and $\vy\tsource \in \Y\tsource$ where we assume $K$-dimensional one-hot encoded class labels, \ie, $\Y\tsource = \mathbb{B}^K$. 
Moreover, we have $n\ttarget$ unlabelled target observations $\data\ttarget = \{\vx\ttarget_j\}^{n\ttarget}_{j=1}$, where $\vx\ttarget \in \X\ttarget$ are $D$-dimensional unlabelled inputs. As in any DA scenario, the assumption made is that the marginal distributions of the source and the target are different, but the semantic concept represented through class labels does not change. 
Formally, we assume that $p(\vy\tsource \mid \vx\tsource) \approx p(\vy\tsource \mid \vx\ttarget)$ and $p(\vx\tsource) \neq p(\vx\ttarget)$. 
In the SFDA scenario we further assume that the source data set is only available while learning the source function $f \colon \X\tsource \to \Y\tsource$ and becomes unavailable while adapting on the unlabelled data. The goal of SFDA is to adapt the source function $f$ to the target domain solely by using the data in $\data\ttarget$. The resulting target function, denoted as $f' \colon \X\ttarget \to \Y\ttarget$, can then be used to infer the class assignment for $\vx\ttarget \in \X\ttarget$. In this work we have considered two settings of the SFDA: i) \emph{vanilla closed-set} SFDA where the label space of the source $\source$ and the target $\target$ is the same, $\labels\tsource = \labels\ttarget$; and ii) \emph{open-set} SFDA where the label space of the $\source$ is a subset of the $\target$, \ie, $\labels\tsource \subset \labels\ttarget$, and $\labels\ttarget \setminus \labels\tsource$ are denoted as \textit{target-private} or OOD classes.

We model the source and target functions $f$ with a neural network that is composed of two sub-networks: feature extractor $g$ and hypothesis function $h$, such that $f = h \circ g$. The feature extractor $g$ and the hypothesis function $h$ are parameterized by parameters $\beta$ and $\theta$, respectively. During target adaptation, the model is initialized with parameters learned on $\data\tsource$ and subsequently the feature extractor parameters are updated using backpropagation, \ie, the hypothesis function is kept frozen.

\begin{figure*}[!t]
  \centering
  \input{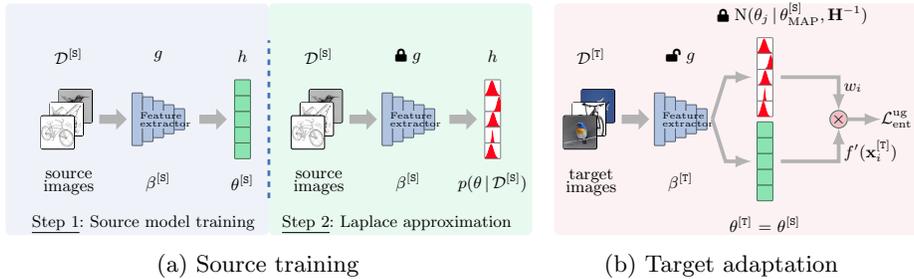}%
  \caption{The pipeline for \ours: (a)~Initial source model training (1) and the additional step (2) of composing a Laplace approximation for assessing the posterior over model parameters, $p(\theta \mid \data\tsource)$. (b)~At target adaptation, we keep the posterior over the parameters fixed ({\scriptsize \faLock}) and train $g$ under a uncertainty-aware composite loss that weights samples according to predictive uncertainty}
  \label{fig:pipeline}
\end{figure*}

\paragraph{Overall Idea} Our proposed method for SFDA operates in two stages. We begin the first stage (see \cref{fig:source-training}) by training a source model on the data set $\data\tsource$, which gives us the maximum-a-posteriori probability (MAP) estimate of the source network parameters ($\{\beta\tsource_\MAP, \theta\tsource_\MAP\}$). 
The second stage (see \cref{fig:target-training}) comprises of maximization of mutual information \cite{gomes2010discriminative} in the predictions for the target inputs $\data\ttarget$. However, due to the overconfidence of ReLU networks \cite{hein2019relu}, maximizing mutual information for all inputs equally, including those that are far away from the source data, could be detrimental. To overcome this pathology, we derive a per-sample weight using the model's uncertainty and use it to modulate the mutual information objective in SHOT.
To estimate the uncertainty in the predictions on the target data, we perform approximate posterior inference over the parameters of the hypothesis function, \ie, $p(\theta\tsource \mid \data\tsource)$.
Inspired by recent works on approximate inference in Bayesian neural networks\cite{mackay1992practical,tierney1986accurate,kristiadi2020being}, we propose to estimate the posterior predictive distribution $p(\vy \mid \vx, \data)$ using a Laplace approximation, introducing little computational overhead and without the need for specialized source training.
We briefly describe the preliminaries to our approach in the following section.

\subsection{Preliminaries}
\label{sec:prelim}

Liang \etal~\cite{liang2020we} proposed SHOT (\textbf{S}ource \textbf{H}yp\textbf{O}thesis \textbf{T}ransfer) for the task of SFDA, where the goal is to find a parameterization $\beta\ttarget$ of the  feature extractor $g$ such that the distribution of latent features $\vz\ttarget = g_{\beta\ttarget}(\vx\ttarget)$ matches the distribution of the latent source features.
This enables that the target data can be accurately classified by the hypothesis function parameterized by $\theta\tsource$. 
To this end, the authors address the SFDA task in two stages where the first and second stage comprise of source model training and maximizing the mutual information \cite{gomes2010discriminative} between the latent  representations and the classifier output, respectively.

The source model $f \colon \X\tsource \to \Y\ttarget$ for a $K$-way classification task is learned using a label-smoothed cross-entropy objective \cite{muller2019does}, \ie,

\begin{equation}
\label{eqn:ce_loss}\textstyle
    \mathcal{L}_{\mathrm{src}} = - \E_{p(\vx\tsource, \vy\tsource)} \sum^{K}_{k=1} \tilde{y}\tsource_{k} \log \phi_k(f(\vx\tsource)),
\end{equation}
where $\phi_k(\va) = \nicefrac{\exp(a_k)}{\sum_j \exp(a_j)}$ denotes the likelihood for the $k$\ths component of the model output and $\tilde{y}\tsource_{i,k} = y\tsource_{i,k} (1 - \alpha) + \nicefrac{\alpha}{K}$  the class label for the $i$\ths label smoothed datum.

After the source training, the $\data\tsource$ is discarded and the target adaptation is conducted on $\data\ttarget$ only. To adapt on the target domain, the target function $f'$ is initialized based on the learned source function $f$ and learned with the information maximization (IM) loss \cite{gomes2010discriminative}.
The IM loss ensures that the function mapping will produce one-hot predictions while at the same time enforcing diverse assignments, \ie,
\begin{align}
    \mathcal{L}_{\mathrm{ent}} &{=}  {-}{\E}_{p(\vx\ttarget)} \textstyle\sum^{K}_{k=1} \phi_k(f'(\vx\ttarget)) \log \phi_k(f'(\vx\ttarget)), \label{eqn:shot-ent} \\
    \mathcal{L}_{\mathrm{div}} &{=} \mathrm{D}_\mathrm{KL}(\hat{\vp} \,\|\, K^{-1} \bm{1}_K) - \log K, \label{eqn:shot-div}
\end{align}
where $\bm{1}_K$ is a vector of all ones, and $\hat{p}_k = \E_{p(\vx\ttarget)}[\phi_k(f'(\vx\ttarget))]$ is the expected network output for the $k$\ths class. Intuitively, $\mathcal{L}_{\mathrm{ent}}$ is in charge of making the network output one-hot, while $\mathcal{L}_{\mathrm{div}}$ is responsible for equally partitioning the network prediction into $K$ classes. In practice $\mathcal{L}_{\mathrm{div}}$ operates on a mini-batch level. In this work we start from SHOT-IM to adapt to the target domain.

\subsection{Uncertainty-guided Source-free DA}
\label{sec:usfan-main-method}

Distributional shift between source and target data sets causes the network outputs to differ, even for the same underlying semantic concept \cite{csurka2017comprehensive}. In a standard UDA scenario, where the source data is available during target adaptation, it is still possible to align the marginal distributions by using a quantifiable discrepancy metric. The task becomes more challenging in the SFDA scenario because it is not possible to align the target feature distribution to a reference (or source) distribution. Moreover, standard ReLU networks are known to yield overconfident predictions for data points which lie far away from the training (source) data \cite{hein2019relu}. In other words, the MAP estimates of a neural network has no notion of uncertainty over the learned weights. Thus, blindly trusting the source model predictions for $\vx\ttarget \in \data\ttarget$ while performing information maximization \cite{liang2020we} or entropy minimization \cite{grandvalet2005semi} can potentially lead to misalignment of clusters between the source and target.

In this work we propose to incorporate the uncertainty of the neural network's weights into the predictions. This mandates a Bayesian treatment of the networks parameters ($\theta$), which gives a posterior distribution over the model parameters by conditioning onto observed data ($\data$), \ie, $p(\theta \mid \data) = \frac{p(\theta)\, p(\data \mid \theta)}{p(\data)} \propto p(\theta)\, p(\data \mid \theta)$.
The prediction of the network $h_{\theta}$ for an observation $\vx$ is given by the predictive posterior distribution, \ie,
\begin{equation}
\label{eqn:predictive}
    p(y_k \mid \vx, \mathcal{D}) = \int_\theta \phi_k(h_{\theta}(\vx)) \, p(\theta \mid \data) \, \mathrm{d}\theta.
\end{equation}
Note that the posterior $p(\theta \mid \data)$ in \cref{eqn:predictive} does not have an analytical solution in general and need to be approximated. For this, we employ a local approximation to the posterior using a Laplace approximation (LA, \cite{tierney1986accurate}). The LA locally approximates the true posterior using a multivariate Gaussian distribution centred at a local maximum and with covariance matrix given by the inverse of the Hessian $\MH$ of the negative log-posterior, \ie, $p(\theta \mid \data) \approx \mathrm{N}(\theta \mid \theta_\MAP, \inv{\MH})$ with $\MH \coloneqq -\nabla^2_{\theta} \log p(\theta \mid \data) \mid_{\theta_\MAP}$. Details can be found in the appendix. Note that the LA is a principled and simple, yet effective, approach to approximate posterior inference stemming from a second-order Taylor expansion of the true posterior around $\theta_\MAP$.  Next we will discuss LA in the context of SFDA.

\paragraph{Bayesian Source Model Generation} In the source training stage (see \cref{fig:source-training}), by optimizing \cref{eqn:ce_loss}, we obtain a MAP estimate of the weights for our source model, comprising $\beta_\MAP$ and $\theta_\MAP$ for $g$ and $h$, respectively. Since $f$ is often modelled by a very deep neural network (\eg, ResNet-50), computing the Hessian can be computationally infeasible owing to the large number of parameters. So we make another simplification by applying a Bayesian treatment only to hypothesis function $h$, known as the \emph{last-layer} Laplace approximation \cite{kristiadi2020being}. This gives us a probabilistic source hypothesis with posterior distribution $p(\theta \mid \data\tsource)$ for the parameters. The feature extractor $g$ remains deterministic. Formally, let $\vz = g_{\beta\tsource}(\vx)$ be the latent feature representation from the feature extractor. Following \cref{eqn:predictive}, the predictive posterior distribution is given as:
\begin{equation}
\label{eqn:last_layer_predictive}
    p(y_k \mid \vz, \data\tsource) \approx \int_\theta \phi_k(h_{\theta}(\vz)) \, \mathrm{N}(\theta \mid \theta_\mathrm{MAP}, \inv{\MH}) \,\mathrm{d}\theta.
\end{equation}
While the \emph{last-layer} LA greatly simplifies the computational overhead for large networks, the Hessian can still be difficult to compute in the case the number of classes is large. To simplify computations, we assume that $\MH$ can be Kronecker-factored $\MH \coloneqq \MV \otimes \MU$ and the resulting approximation is referred to as Kronecker-factored Laplace approximation (KFLA, \cite{ritter2018scalable}). Such probabilistic treatment allows us to quantify uncertainty in the predictions for data points from the target with little computational overhead. Also, the LA can be readily computed using a single forward pass of the source data through the network. Next, we describe how to use the uncertainty estimates during target adaptation.

\paragraph{Uncertainty-guided Information Maximization}
Upon completion of the source model generation stage, we exploit the probabilistic source hypothesis to guide the information maximization in the target adaptation stage. SHOT puts equal confidence on all the target predictions and do not make any distinction for the target feature that lies outside of the source manifold. We emphasize that in case of strong domain-shift na\"ively maximizing the IM loss could lead to cluster misalignment. For that reason, we propose to weigh the entropy minimization objective (\cref{eqn:shot-ent}) with a weight which is proportional to the certainty in the target predictions (see \cref{fig:target-training}). To get the per-sample weight for a $\vx\ttarget$ we need to compute the predictive posterior distribution, as outlined in \cref{eqn:last_layer_predictive}. However, exactly solving the integration is intractable in many cases and we, therefore, resort to Monte Carlo (MC) integration.
Let $\vz\ttarget = g_{\beta\ttarget}(\vx\ttarget)$, the approximate predictive posterior distributions is:
\begin{equation}
\label{eqn:last_layer_target}
    p(y_k \mid \vz\ttarget, \data\tsource) \approx \frac{1}{M} \sum^{M}_{j=1} \phi_k\left(h_{\theta_j}(\vz\ttarget)\right),
\end{equation}
where $\theta_j \sim \mathrm{N}(\theta_j \mid \theta_{\MAP}, \inv{\MH})$ and $M$ denotes the number of MC steps. To encourage low entropy predictions we additionally scale the outputs of the hypothesis by $1/\tau$, where $0<\tau\leq 1$. 
The final weight of each observation $\vx_i\ttarget$ is then computed as $w_i = \exp(-H)$ where $H$ denotes the entropy of the predictive mean. The \emph{uncertainty-guided entropy loss} is then given as:
\begin{equation}
\label{eqn:uncertainty_shot}
\mathcal{L}^\mathrm{ug}_\mathrm{ent} = {-} {\E}_{p(\vx\ttarget)} \sum^{K}_{k=1} w \, \sigma_k(f'(\vx\ttarget))\log\sigma_k(f'(\vx\ttarget)).
\end{equation}
The final training objective is then given as: 
    $\mathcal{L}_\text{U-SFAN} = (1-\gamma)\mathcal{L}^\mathrm{ug}_\mathrm{ent} + \gamma \, \mathcal{L}_\mathrm{div}$.
Pseudocode for our \ours can be found in the appendix.

\paragraph{How does this differ from conventional uncertainty estimation?} The importance and advantages of adopting a Laplace approximation (LA) over Monte Carlo (MC) dropout to estimate uncertainty in SFDA can be summarized as follows: {\em (i)} LA does not require specialized network architecture (\eg, dropout layers), loss function, or re-training (as in MC dropout) to estimate predictive uncertainties. This greatly decouples the source training from target adaptation, which is essential to be applicable in SFDA; {\em (ii)} To have well-calibrated uncertainties, MC dropout requires a grid search over the dropout probabilities \cite{gal2017concrete}, a prohibitive operation in deep neural networks, especially as the future target data is not available at source training. LA is a more principled approach that does not require a grid search, making it better suited for SFDA. {\em (iii)} LA is computationally lightweight since it requires just a single forward pass of the source data through the network after the source training to estimate the posterior over the parameters of the sub-network. {\em (iv)} LA does not impact the training time during target adaptation because, unlike MC dropout, only a single forward pass is needed to quantify the predictive uncertainties. Because LA employs a Gaussian approximation to the posterior, MC integration is cheap and efficient to compute. {\em (v)} As used in our work, LA estimates the full posterior over the weights and biases, while MC dropout can only account for the uncertainties over the weights \cite{gal2016dropout} and is known to be a poor approximation to the posterior \cite{osband2016risk,osband2018randomized,foong2020expressiveness}. {\em (vi)} LA preserves the decision boundary induced by the MAP estimate, which is not the case for MC dropout \cite{kristiadi2020being}. In summary, our contribution goes beyond the uncertainty re-weighting scheme \cite{liang2020balanced,liang2019exploring} commonly used in UDA, while carrying many advantages over existing works.


\section{Experiments}
\label{sec:exp}

We conduct experiments on four standard DA benchmarks: {\sc Office31} \cite{saenko2010adapting}, {\sc Office-Home} \cite{venkateswara2017deep}, {\sc Visda-C} \cite{peng2017visda}, and the large-scale {\sc DomainNet} \cite{peng2019moment} (\textbf{0.6 million} images). The details of the benchmarks are summarized in appendix. For the experiments in the open-set DA setting we follow the split of \cite{liang2020we} for shared and target-private classes.

\paragraph{Evaluation protocol} We report the classification accuracy for every possible pair of $source \mapsto target$ directions, except for the {\sc Visda-C} where we are only concerned with the transfer from \textit{synthetic} $\mapsto$ \textit{real} domain. For the open-set experiments, following the evaluation protocol in \cite{liang2020we}, we report the OS accuracy which includes the per-class accuracy of the known and the unknown class and is computed as $\mathrm{OS} = \frac{1}{K+1} \sum^{K+1}_{k=1}\mathrm{acc}_k$, where $k=\{1, 2, \dots, K\}$ denote the shared classes and $(K+1)$\ths is the target-private or OOD classes. This metric is preferred over the known class accuracy, $\mathrm{OS}^*=\frac{1}{K}\sum^{K}_{k=1}\mathrm{acc}_k$, as it does not take into account the OOD classes.

\paragraph{Implementation details} We adopted the network architectures used in the SFDA literature, which are ResNet-50 or ResNet-101 \cite{he2016deep}. Following \cite{liang2020we}, we added a bottleneck layer containing 256 neurons which is then followed by a batch normalization layer. The network finally ends with a weight normalized linear classifier that is kept frozen during the target adaptation. Details about hyperparameters can be found in the appendix. For computing the KFLA we use the PyTorch package of Dangel \etal \cite{dangel2019backpack}. Our code is available at \url{https://github.com/roysubhankar/uncertainty-sfda}

\subsection{Ablation Studies}
\label{sec:abl}
\begin{figure}[t!]
\scriptsize\centering
\setlength{\figurewidth}{.14\textwidth}
\captionsetup[subfigure]{justification=centering}
\tikzstyle{label} = [rotate=90,align=center,text width=.9\figurewidth,minimum height=4mm]

\begin{minipage}{.1\textwidth}
~
\end{minipage}
\begin{minipage}{\figurewidth}
  \centering
  \tikz\node[]{\sc Source};
\end{minipage}
\begin{minipage}{\figurewidth}
  \centering
  \tikz\node[]{\sc Target};
\end{minipage}
\begin{minipage}{\figurewidth}
  \centering
  \tikz\node[]{\sc SFDA (IM)};
\end{minipage}
\hfill
\begin{minipage}{\figurewidth}
  \centering
  \tikz\node[]{\sc Source};
\end{minipage}
\begin{minipage}{\figurewidth}
  \centering
  \tikz\node[]{\sc Target};
\end{minipage}
\begin{minipage}{\figurewidth}
  \centering
  \tikz\node[]{\sc SFDA (IM)};
\end{minipage}
\\
\begin{minipage}{.1\textwidth}
  \raggedleft
  \tikz\node[label]{{\sc ~ MAP} \mbox{\tiny (conventional)}};\\
  \tikz\node[label]{\sc Uncertainty-guided};%
\end{minipage}
\begin{subfigure}{3.1\figurewidth}
  \raggedleft
  \includegraphics[width=\figurewidth]{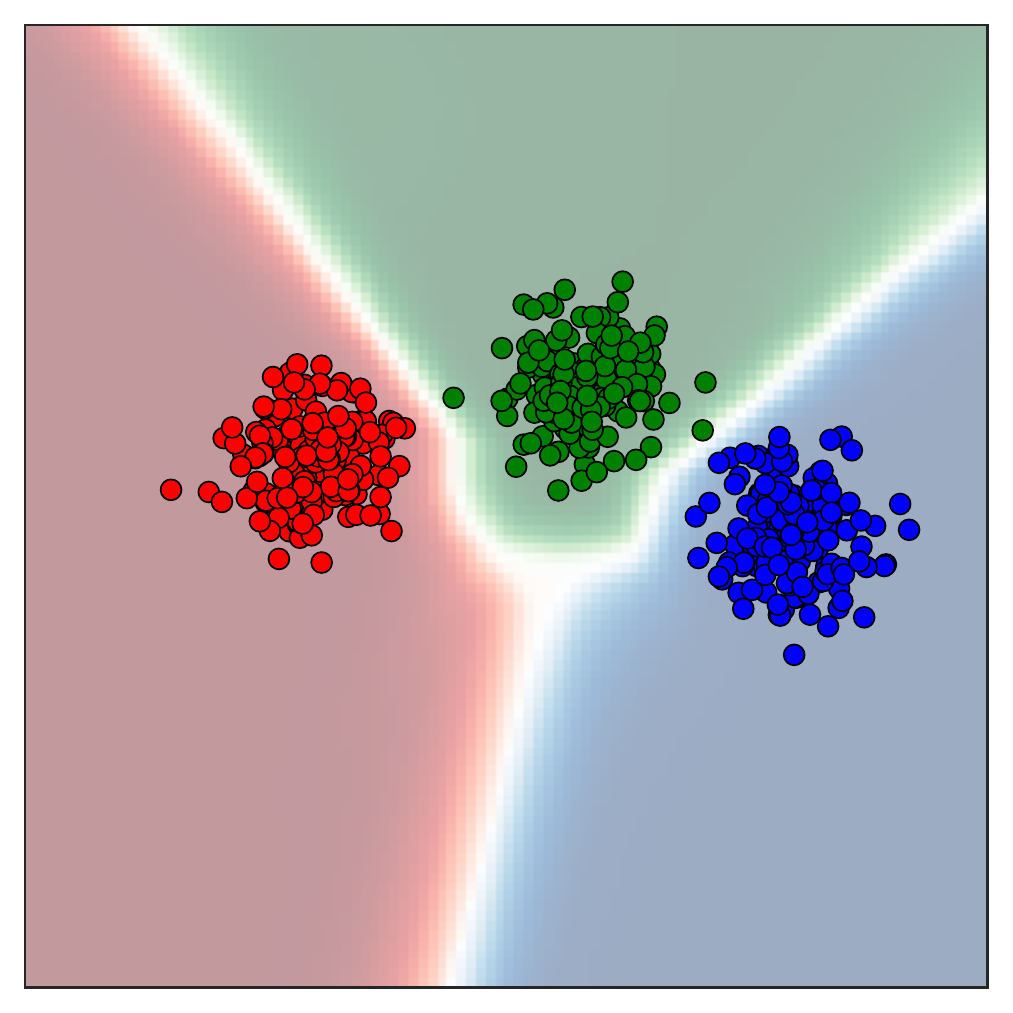}
  \includegraphics[width=\figurewidth]{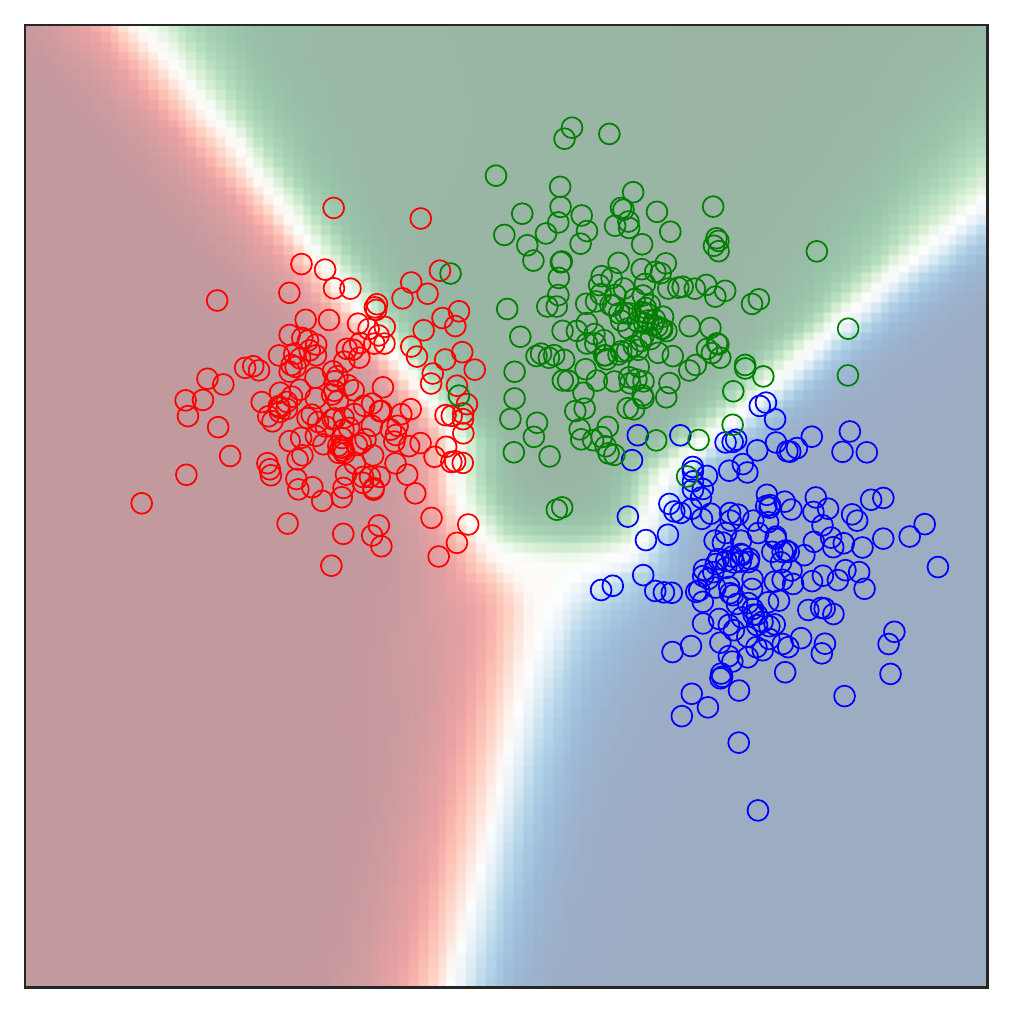}
  \includegraphics[width=\figurewidth]{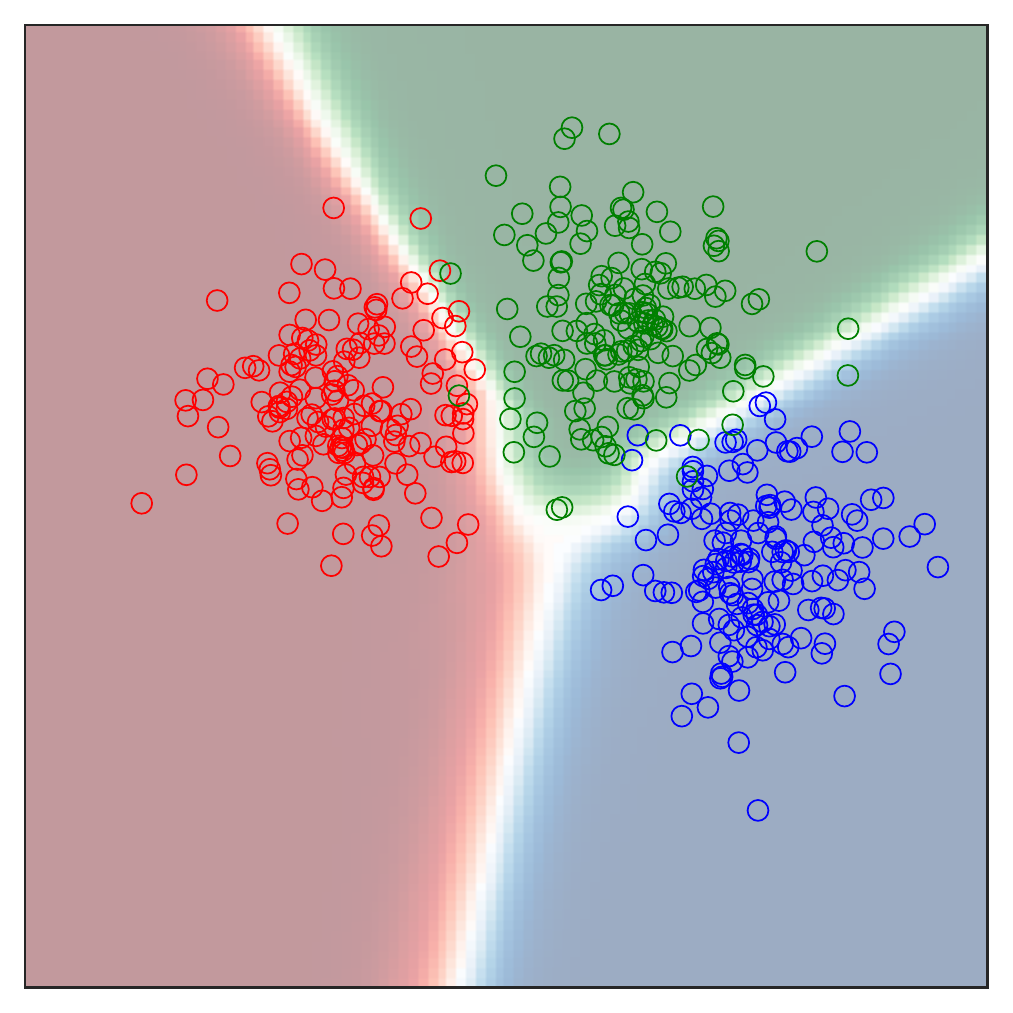}
  \\
  \includegraphics[width=\figurewidth]{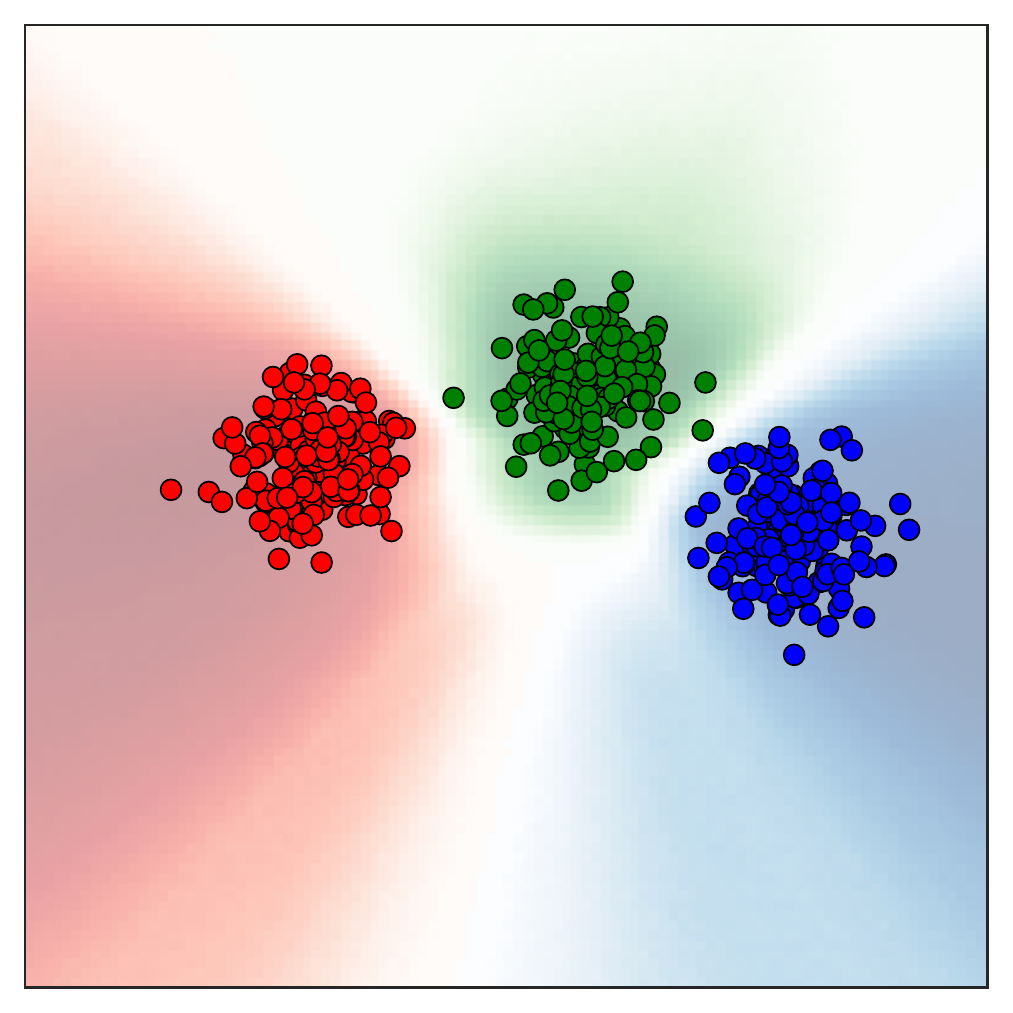}
  \includegraphics[width=\figurewidth]{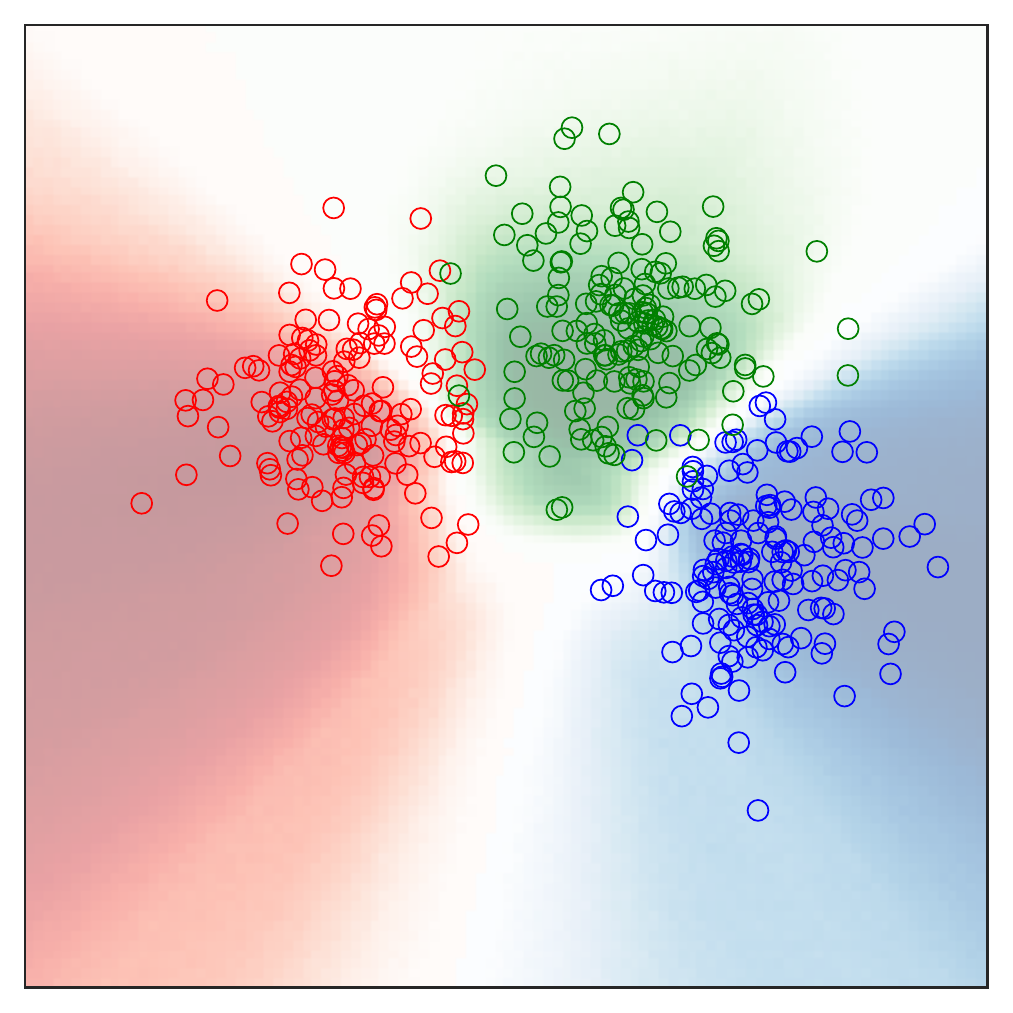}
  \includegraphics[width=\figurewidth]{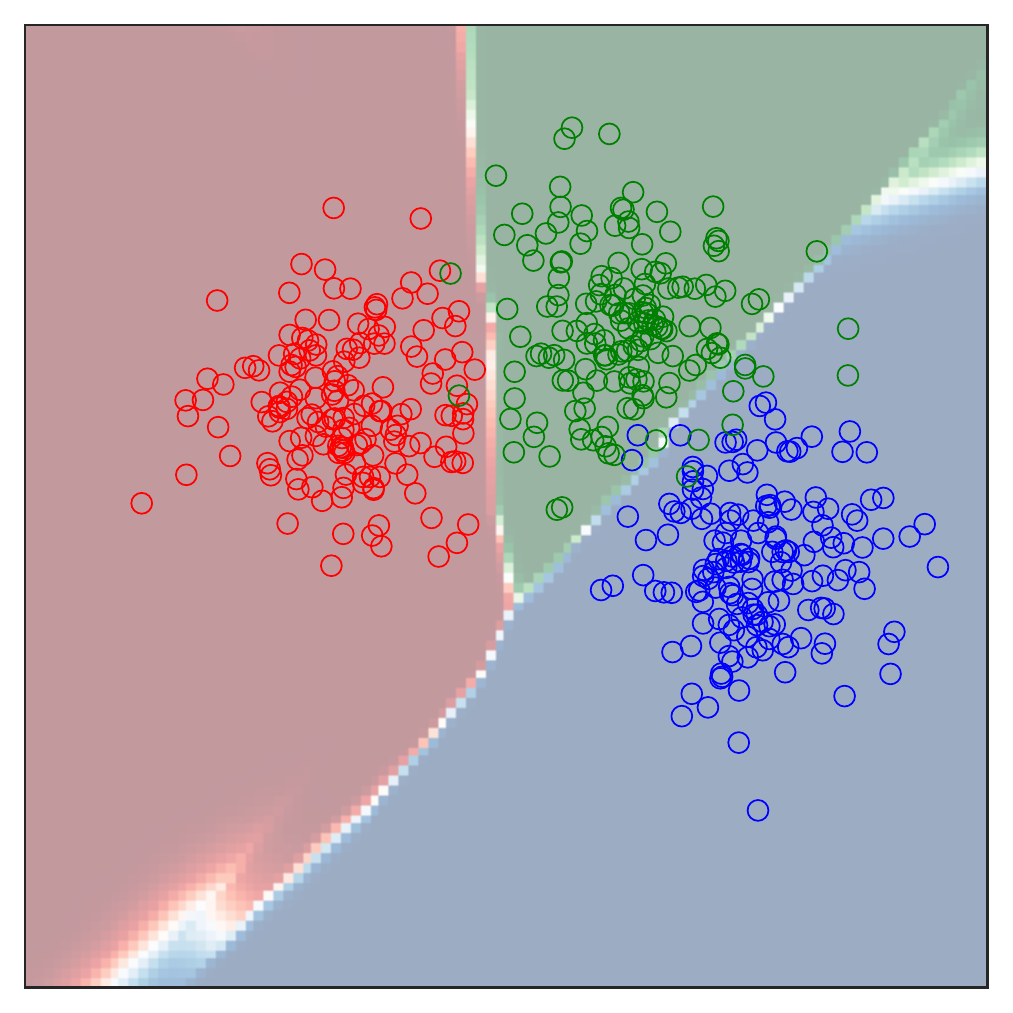}
  \caption{Mild domain shift}
  \label{fig:toy_experiment-mild}
\end{subfigure}
\hfill
\begin{subfigure}{3.1\figurewidth}
  \raggedleft
  \includegraphics[width=\figurewidth]{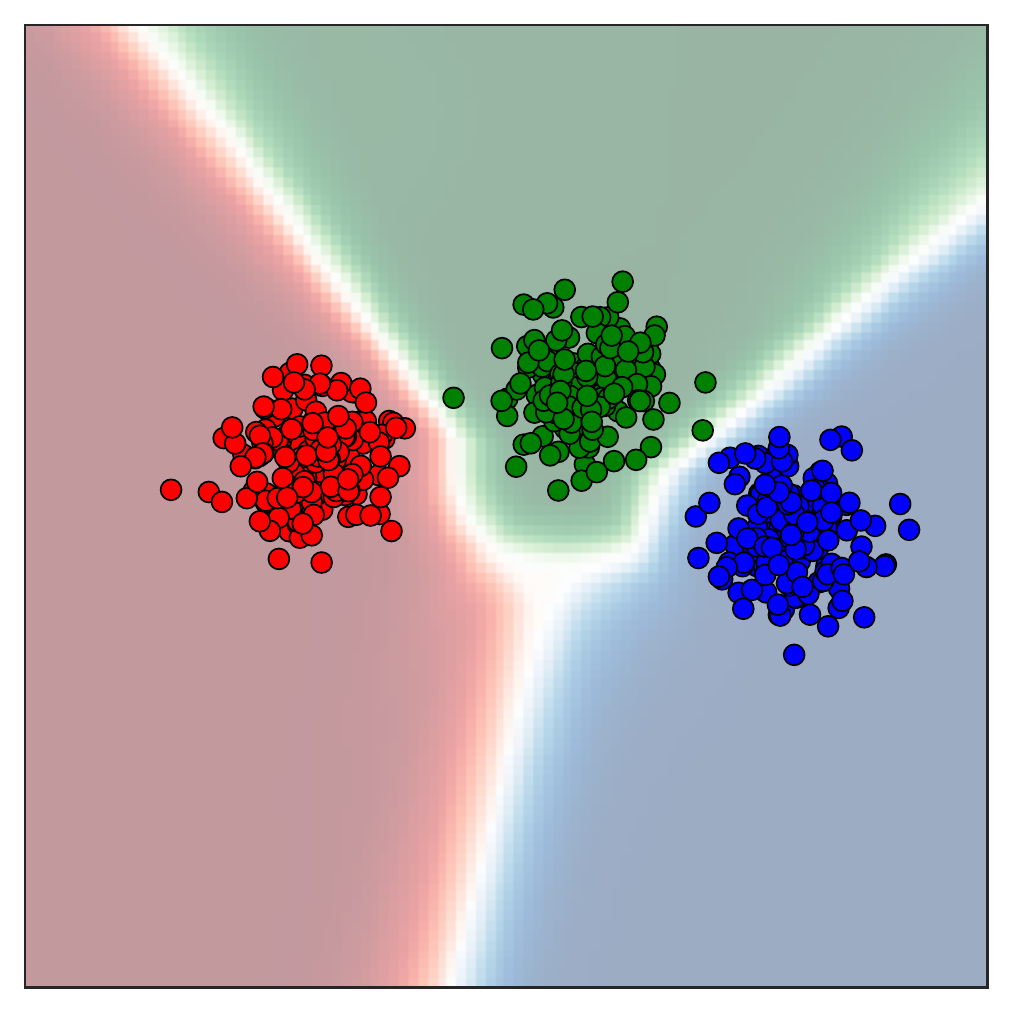}
  \includegraphics[width=\figurewidth]{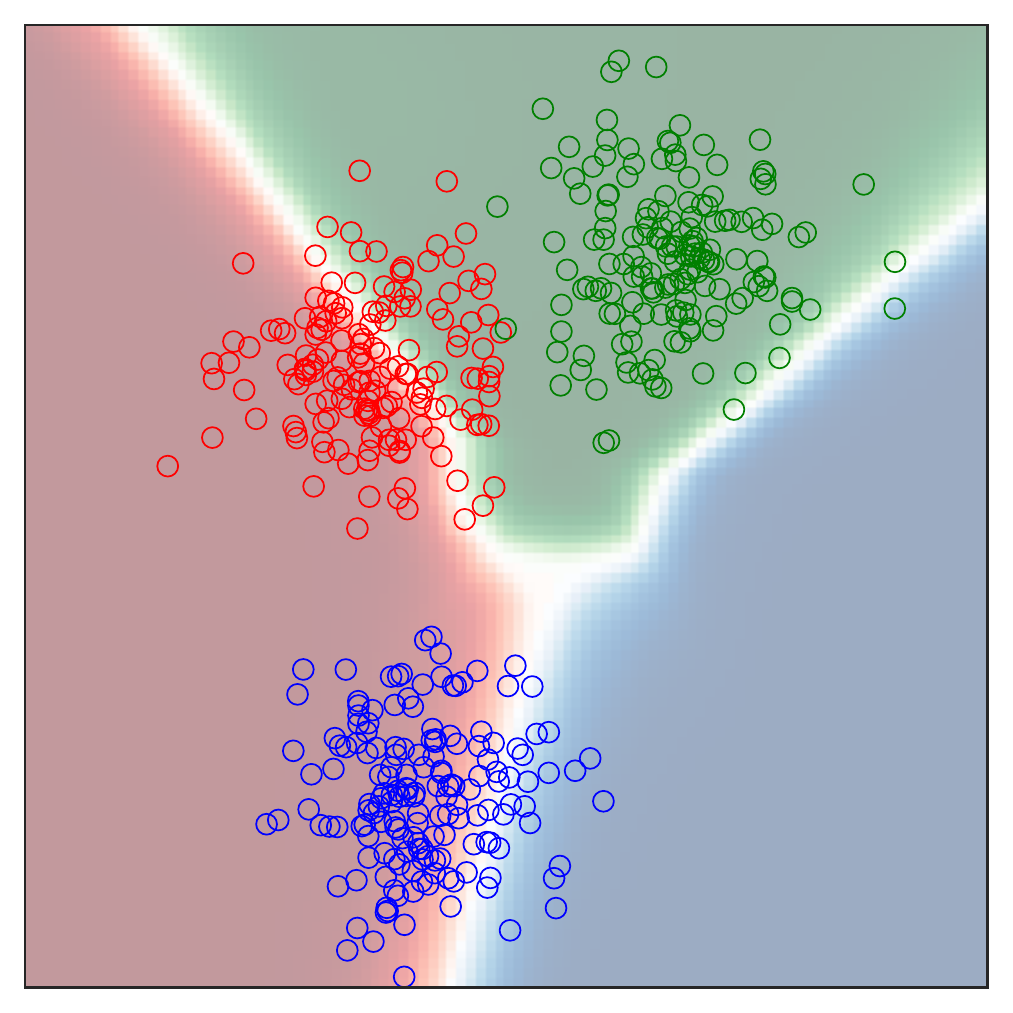}
  \includegraphics[width=\figurewidth]{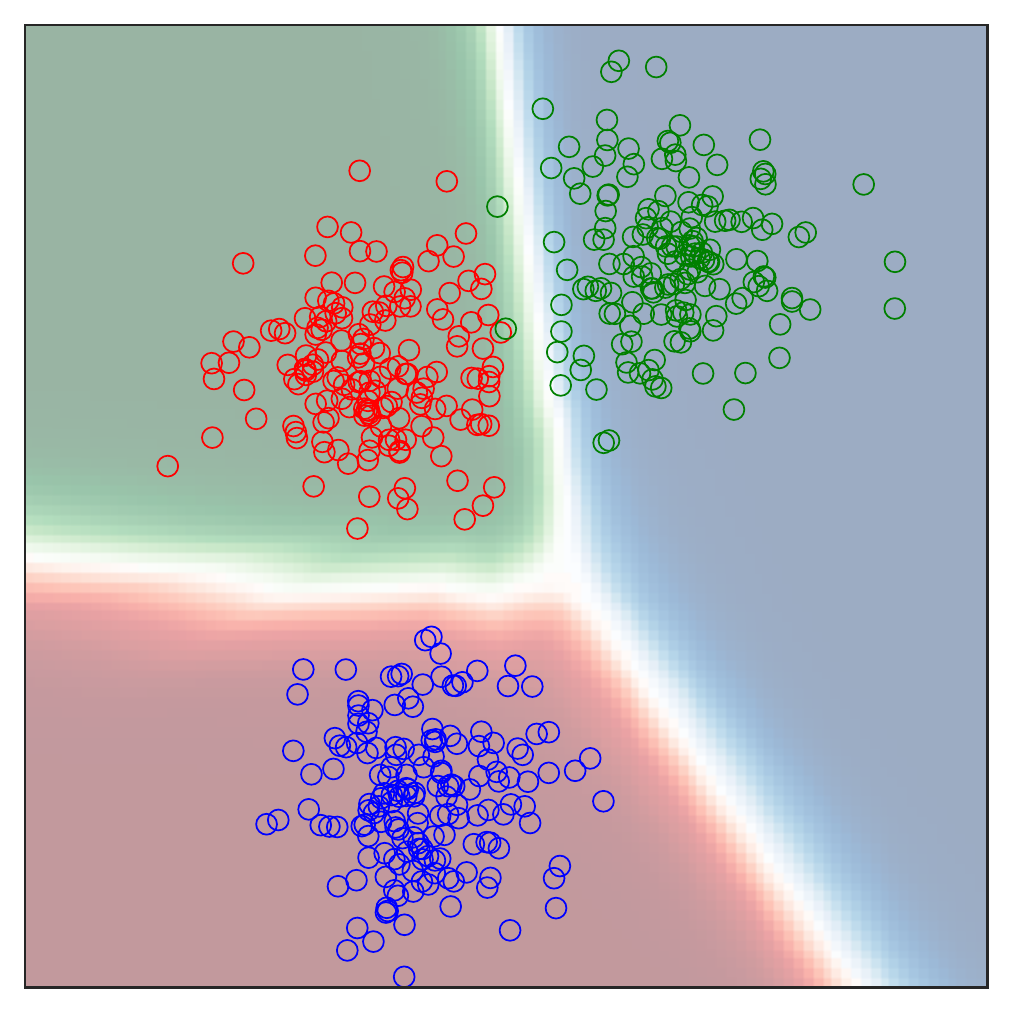}
  \\
  \includegraphics[width=\figurewidth]{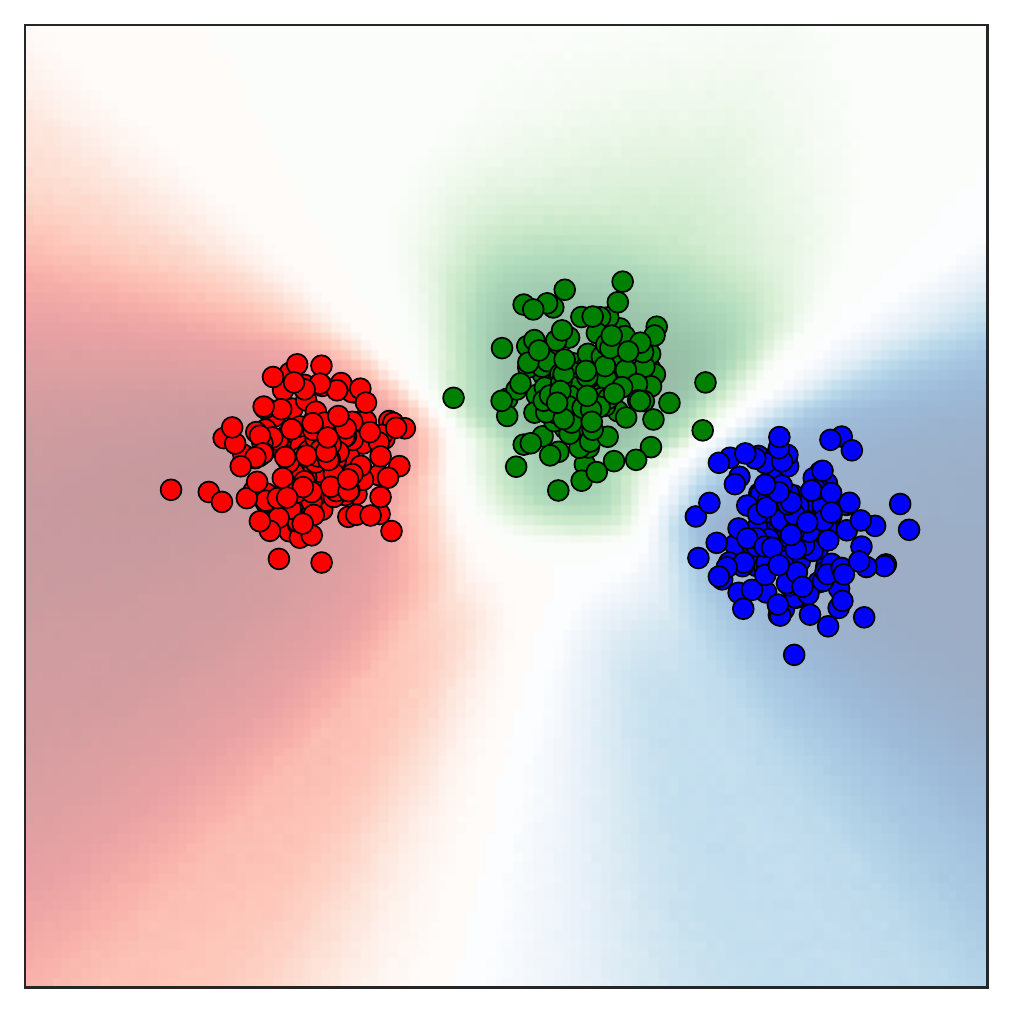}
  \includegraphics[width=\figurewidth]{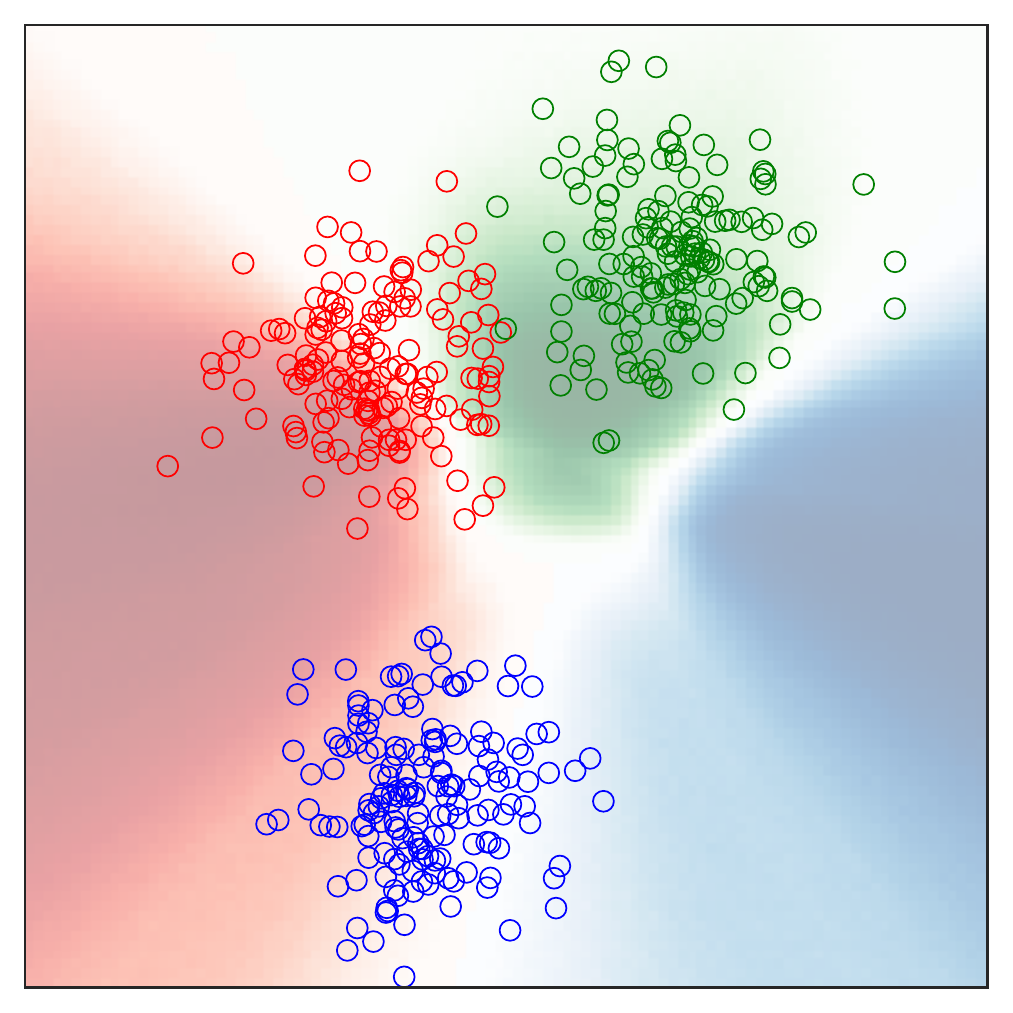}
  \includegraphics[width=\figurewidth]{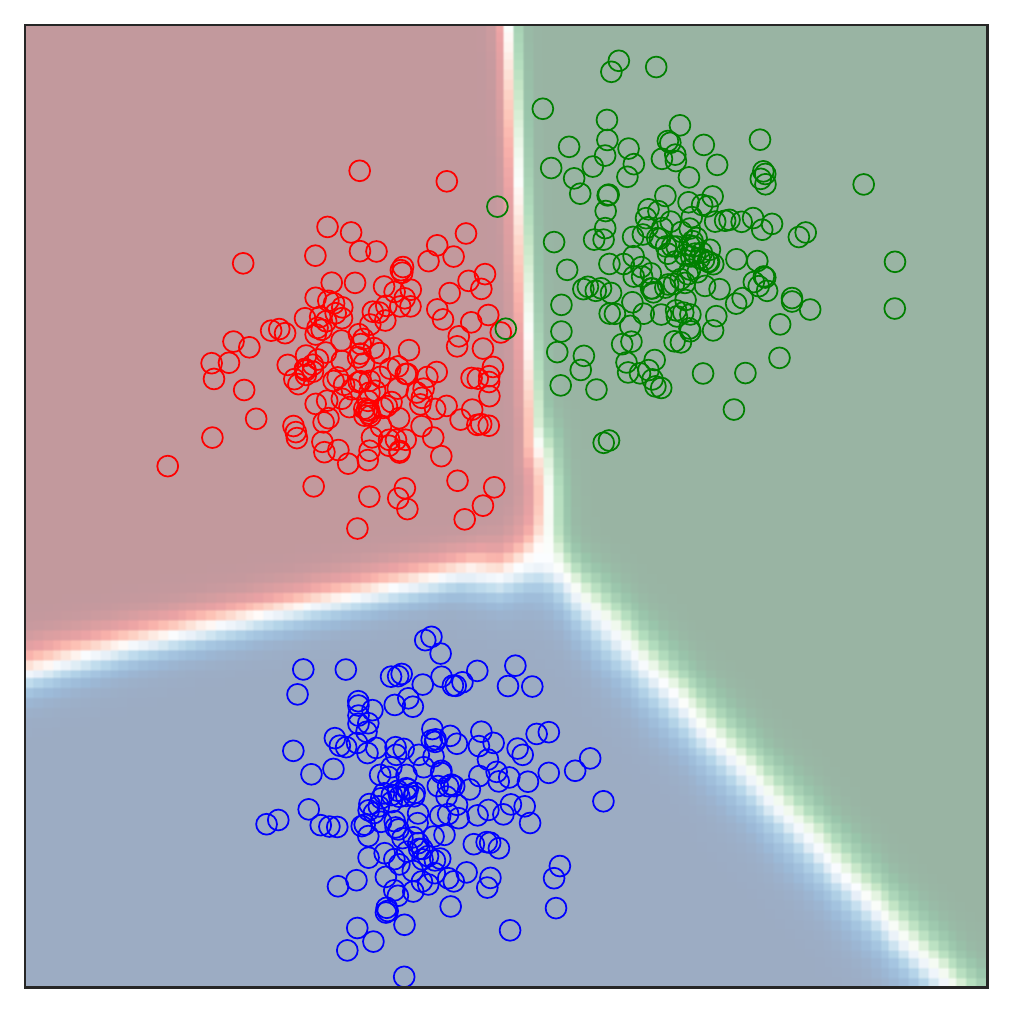}
  \caption{Strong domain shift}
  \label{fig:toy_experiment-strong}
\end{subfigure}
\caption{Comparison of conventional IM (MAP) with our uncertainty-guided IM on target data under mild and strong domain-shift. The solid \protect\tikz[baseline=-.6ex]\protect\node[circle,draw=black,fill=black,inner sep=1pt]{}; vs.\ hollow \protect\tikz[baseline=-.6ex]\protect\node[circle,draw=black,fill=white,inner sep=1pt]{}; circles represent the source and the target data, respectively. Each class is colour coded and the decision boundaries are shaded with the corresponding colours. Under strong domain-shift, IM, when used with a MAP estimate, finds a {\bf completely flipped} decision boundary. \ours finds the decision boundary by down-weighting the \emph{far away} target data}
\label{fig:toy_experiment}
\end{figure}

As discussed in \cref{sec:usfan-main-method}, conventional SFDA methods that rely on optimizing the IM loss on the unlabelled target data (\eg, SHOT) are prone to misalignment of the target data with the source hypothesis under strong domain shift. To visually demonstrate this phenomenon, we design an experiment of a 3-way classification task on toy data (see \cref{fig:toy_experiment}). Given a set of source data points, belonging to three classes, we simulate two kinds of domain-shift: \emph{mild} shift (\cref{fig:toy_experiment-mild}) and \emph{strong} shift (\cref{fig:toy_experiment-strong}). In the case of mild shift, the target data points stay very close to the source manifold, and the conventional approach (only using the MAP estimate) can classify a majority of target data points without the need of adaptation. Whereas, in the case of strong shift, the target data points for the blue class, in particular, shift drastically away from the source points. The source model based on the MAP estimate misclassifies most of the target data points with high confidence. On the other hand, our uncertainty-guided source model remains certain only for those target points which lie within the source support and assigns low certainty otherwise (proportional to the strength of colours depicting the decision surface in \cref{fig:toy_experiment}), robustifying the adaptation on the target data in case of domain shift.

Given such a set-up, we optimize the IM loss (\ie, SHOT-IM) for both the conventional and the uncertainty-guided source models. In the case of mild shift, both can reliably partition the target data points under the right decision surfaces (see \cref{fig:toy_experiment-mild} (right)). This is intuitive because the decision boundary of the target model already passes through the low-density regions. Hence, the optimization of the IM loss leads to correct target classification with both methods. However, when the domain shift is more substantial, the conventional approach results in \emph{completely flipped} decision boundaries. This happens because most blue target points fall under the red decision surface, and thus, the IM loss assigns them to class `red'. On the contrary, our uncertainty-guided approach down-weights the blue points, and \emph{safely} optimizes the IM loss as the model is uncertain about the class assignment for those points (\cref{fig:toy_experiment-strong} (right)). This protects from major changes in the decision boundaries and allows the optimization to find the correct decision boundaries for the target data. Therefore, highlighting the importance of having a notion of uncertainty in the model predictions during adaptation. We show later that this intuition also holds well for real-world data sets where our \ours offers more robustness when the domain-shift in a data set becomes challenging. 

\begin{figure}[h!]
    \centering
    \pgfplotsset{axis on top,scale only axis,y tick label style={rotate=90}} 
    \setlength{\figurewidth}{.4\columnwidth}
    \setlength{\figureheight}{0.67\figurewidth}
    \begin{subfigure}[t]{0.4\columnwidth}
        \centering\scriptsize
        \pgfplotsset{ylabel={Density (arbitrary units)}}
        \input{figs/cifar/CIFAR9_STL9_displot_entropy_before}%
        \caption{Before adaption}
        \label{fig:cifar_stl_entropy-a}
    \end{subfigure}
    \hfill
    \begin{subfigure}[t]{0.4\columnwidth}
        \centering\scriptsize
        \input{figs/cifar/CIFAR9_STL9_displot_entropy_after}%
        \caption{After adaption} 
        \label{fig:cifar_stl_entropy-b}       
    \end{subfigure}
    \caption{Entropy density plots for CIFAR9 $\rightarrow$ STL9 in the closed-set SFDA setting using the MAP estimate (\protect\tikz[baseline=-0.5ex]\protect\draw[dashed]{(0,0) -- (0.55,0)};~correct, \protect\tikz[baseline=-0.5ex]\protect\draw[red,dashed] {(0,0) -- (0.55,0)};~incorrect) or our approach (\protect\tikz[baseline=-0.5ex]\protect\draw[thick]{(0,0) -- (0.55,0)};~correct,\protect\tikz[baseline=-0.5ex]\protect\draw[red, thick]{(0,0) -- (0.55,0)};~incorrect). Our uncertainty-guided SFDA approach places less mass on low-entropy incorrect samples before and after adaptation}
    \label{fig:cifar_stl_entropy}
\end{figure}

To gain further insights, we visualize the entropy density plots of the source model predictions before and after adaptation with conventional (MAP estimate) and uncertainty-guided models on an image data set (CIFAR \cite{krizhevsky2009learning} as source data set and STL \cite{coates2011analysis} as target). As shown in \cref{fig:cifar_stl_entropy-a}, the MAP estimate has lower entropy predictions for both the correct and incorrect predictions, when compared to our uncertainty-guided model. Reduced over-confidence for our approach is expected before the adaptation phase, however, it is non-trivial that this behavior also bears in the post-adaptation phase. The reduced over-confident allows our \ours to down-weight the incorrect predictions during target adaptation, resulting in improved target accuracy over SHOT-IM ($77.04\%$ for \ours vs $75.69\%$ for SHOT-IM). This effect can be noticed in \cref{fig:cifar_stl_entropy-b} where \ours has overall higher entropy incorrect predictions, which is desirable in SFDA.

\begin{table}[!t]
  \centering
  \caption{Comparison of model performance using entropy weighting during target adaptation on the {\sc Office-Home} data set. The weights computed using the LA is more beneficial than the weights computed with a MAP network}
  \scriptsize
  \setlength{\tabcolsep}{7pt}
  \begin{tabularx}{\textwidth}{lcccc}
       \specialrule{1pt}{1pt}{1pt}
       \sc Method & \sc Source-Only & \sc SHOT-IM \cite{liang2020we} & \sc SHOT-IM + Ent.\ weighting & \sc \ours (ours) \\
       \specialrule{1pt}{1pt}{1pt}
       \sc Avg.\ Acc.
        & 60.3
        & 70.5
        & 71.2
        & \textbf{71.8} \\
       \specialrule{1pt}{1pt}{1pt}
  \end{tabularx}
  \label{tab:officehome_ablation}
\end{table}

To further understand the contribution of our uncertainty-guided re-weighting, we run an ablation where the approximate posterior distribution of our method (\cref{eqn:uncertainty_shot}) is replaced by a weight computed from a point estimate from a MAP source model. This model is denoted as SHOT-IM + ENT. WEIGHTING in \cref{tab:officehome_ablation}. We observe that such weighting scheme indeed improves the performance over SHOT-IM. However, it still lacks behind our proposed \ours which uses weights computed from the uncertainty-guided model. This clearly shows that the improvement in performance with \ours is not simply caused by the re-weighting but also due to better identification of target samples that are not well explained under the source model.

\subsection{State-of-the-art Comparison}\label{sec:sota}

\begin{table*}[t!]
    \centering
    \caption{Comparison of the classification accuracy on the {\sc Office-Home} for the closed-set setting using ResNet-50. High overall performance signifies \textit{milder} distributional shift between domains. The improvement of \ours upon SHOT is moderate, but competitive w.r.t. $\text{A}^2\text{Net}$\cite{xia2021adaptive} or SHOT\Plus\Plus~\cite{liang2021source}, which require complex training objectives}
    \scriptsize
    \tableoptions
    \setlength{\tabcolsep}{1.3pt}
    \begin{tabularx}{\textwidth}{lcccccccccccca} 
    \specialrule{1pt}{1pt}{1pt}
    \sc Method & A$\rightarrow$C & A$\rightarrow$P & A$\rightarrow$R & C$\rightarrow$A & C$\rightarrow$P & C$\rightarrow$R & P$\rightarrow$A & P$\rightarrow$C & P$\rightarrow$R & R$\rightarrow$A & R$\rightarrow$C & R$\rightarrow$P & \sc Avg. \\  
    \specialrule{1pt}{1pt}{1pt}
    ResNet-50 & 34.9 & 50.0 & 58.0 & 37.4 & 41.9 & 46.2 & 38.5 & 31.2 & 60.4 & 53.9 & 41.2 & 59.9 & 46.1 \\
    DANN \cite{ganin2016domain} & 45.6 & 59.3 & 70.1 & 47.0 & 58.5 & 60.9 & 46.1 & 43.7 & 68.5 & 63.2 & 51.8 & 76.8 & 57.6 \\
    DWT \cite{roy2019unsupervised} & 50.3 & 72.1 & 77.0 & 59.6 & 69.3 & 70.2 & 58.3 & 48.1 & 77.3 & 69.3 & 53.6 & 82.0 & 65.6 \\
    CDAN \cite{long2018conditional} & 50.7 & 70.6 & 76.0 & 57.6 & 70.0 & 70.0 & 57.4 & 50.9 & 77.3 & 70.9 & 56.7 & 81.6 & 65.8\\
    SAFN \cite{xu2019larger} & 52.0& 71.7& 76.3& 64.2& 69.9& 71.9& 63.7& 51.4& 77.1& 70.9& 57.1& 81.5& 67.3 \\
    \hline
    SHOT-IM \cite{liang2020we} & 55.4& 76.6& 80.4& 66.9& 74.3& 75.4& 65.6& 54.8& 80.7& 73.7& 58.4& 83.4& 70.5 \\
    LSC \cite{yang2021generalized} & 57.9 & 78.6 & 81.0 & 66.7 & 77.2 & 77.2 & 65.6 & 56.0 & 82.2 & 72.0 & 57.8 & 83.4 & 71.3 \\
    \ours (Ours) & 58.5 & 78.6 & 81.1 & 66.6 & 75.2 & 77.9 & 66.3 &	57.9 & 80.6 & 73.6 & 61.4 & 84.1 & 71.8 \\
    $\text{A}^2\text{Net}$\cite{xia2021adaptive} & 58.4 & 79.0 & 82.4 & 67.5 & 79.3 & 78.9 & 68.0 & 56.2 & 82.9 & 74.1 & 60.5 & 85.0 & 72.8 \\
    SHOT\Plus\Plus~\cite{liang2021source} & 57.9 & 79.7 & 82.5 & 68.5 & 79.6 & 79.3 & 68.5 & 57.0 & 83.0 & 73.7 & 60.7 & 84.9 & 73.0 \\
    \hline
    SHOT \cite{liang2020we}  & 57.1& 78.1& 81.5& 68.0& 78.2& 78.1& 67.4& 54.9& 82.2& 73.3& 58.8& 84.3& 71.8 \\
    \ours\Plus\: (Ours) & 57.8 & 77.8	& 81.6 & 67.9 & 77.3 & 79.2 & 67.2 & 54.7 &	81.2 & 73.3	& 60.3 & 83.9 & 71.9\\
    \specialrule{1pt}{1pt}{1pt}
    \end{tabularx}
    \label{tab:cda-office-home}
\end{table*}

\begin{table*}[h!]
    \centering
    \caption{Comparison of the OS classification accuracy on the {\sc Office-Home} for the open-set setting using ResNet-50. \ours improves over SHOT without the need for nearest-centroid pseudo-labelling in the case of open-set SFDA}
    \scriptsize
    \tableoptions
    \setlength{\tabcolsep}{1.3pt}
    \begin{tabularx}{\textwidth}{lcccccccccccca} 
    \specialrule{1pt}{1pt}{1pt}
    \sc Method & A$\rightarrow$C & A$\rightarrow$P & A$\rightarrow$R & C$\rightarrow$A & C$\rightarrow$P & C$\rightarrow$R & P$\rightarrow$A & P$\rightarrow$C & P$\rightarrow$R & R$\rightarrow$A & R$\rightarrow$C & R$\rightarrow$P & \sc Avg. \\ 
    \specialrule{1pt}{1pt}{1pt}
    ResNet-50 & 53.4 & 52.7 & 51.9 & 69.3 & 61.8 & 74.1 & 61.4 & 64.0 & 70.0 & 78.7 & 71.0 & 74.9 & 65.3 \\
    ATI-$\lambda$\cite{panareda2017open} & 55.2 & 52.6 & 53.5 & 69.1 & 63.5 & 74.1 & 61.7 & 64.5 & 70.7 & 79.2 & 72.9 & 75.8 & 66.1 \\
    OpenMax \cite{bendale2016towards} & 56.5 & 52.9 & 53.7 & 69.1 & 64.8 & 74.5 & 64.1 & 64.0 & 71.2 & 80.3 & 73.0 & 76.9 & 66.7 \\
    STA \cite{liu2019separate} & 58.1 & 53.1 & 54.4 & 71.6 & 69.3 & 81.9 & 63.4 & 65.2 & 74.9 & 85.0 & 75.8 & 80.8 & 69.5 \\
    \hline
    SHOT-IM \cite{liang2020we} & 62.5 & 77.8 & 83.9 & 60.9 & 73.4 & 79.4 & 64.7 & 58.7 & 83.1 & 69.1 & 62.0 & 82.1 & 71.5 \\
    SHOT \cite{liang2020we} & 64.5 & 80.4 & 84.7 & 63.1 & 75.4 & 81.2 & 65.3 & 59.3 & 83.3 & 69.6 & 64.6 & 82.3 & 72.8 \\
    \ours(Ours)\phantom{+} & 62.9 & 77.9	& 84.0 & 67.9 & 74.6 & 79.6 & 68.8 & 61.3	& 83.3 & 76.0 & 63.9 & 82.3	& 73.5\\
    \specialrule{1pt}{1pt}{1pt}
    \end{tabularx}
    \label{tab:oda-office-home}
\end{table*}

\begin{table}[!t]
\caption{(a)~Comparison of the classification accuracy on the {\sc Visda-C} for the closed-set DA, pertaining to the \textit{Synthetic $\rightarrow$ Real} direction, using ResNet-101. $\dagger$ indicates the numbers of \cite{liang2020we} that are obtained using the official code from the authors. Note that several SFDA methods perform equally well for {\sc visda-c}, hinting at saturating performance.
(b)~Comparison of the average accuracy on the {\sc Domainnet} for the closed-set SFDA using ResNet-50. The {\sc source} column indicates the domain where the source model has been trained. The data set being challenging (exhibiting \textit{strong} domain-shift), the improvement with our \ours over \cite{liang2020we} is substantial}
\begin{subfigure}[t]{0.48\textwidth}
    \centering
    \caption{{\sc Visda-C}}
    \scriptsize
    \tableoptions
    \setlength{\tabcolsep}{11pt}
    \begin{tabularx}{0.8\columnwidth}{lc} 
    \specialrule{1pt}{1pt}{1pt}
    \sc Method & \sc Acc. \\
    \specialrule{1pt}{1pt}{1pt}
    ResNet-101 & 52.4 \\
    CDAN+BSP \cite{chen2019transferability} & 75.9 \\
    SAFN \cite{xu2019larger} & 76.1 \\
    \hline
    $\text{SHOT-IM}^{\dagger}$ \cite{liang2020we} & 80.3 \\
    \ours (Ours) & 81.2 \\
    3C-GAN \cite{li2020model} & 81.6 \\
    $\text{A}^2\text{Net}$\cite{xia2021adaptive} & 84.3 \\
    \hline
    $\text{SHOT}^{\dagger}$ \cite{liang2020we} & 82.4 \\
    \ours\Plus\: (Ours) & 82.7 \\
    \specialrule{1pt}{1pt}{1pt}
    \end{tabularx}
    
    \label{tab:cda-visda}
\end{subfigure}
\hfill
\begin{subfigure}[t]{0.48\textwidth}
    \centering
    \caption{{\sc Domainnet}}
    \scriptsize
    \tableoptions
    \setlength{\tabcolsep}{9pt}
    \begin{tabularx}{\columnwidth}{lcc}
         \specialrule{1pt}{1pt}{1pt}
         \sc Source & \sc SHOT-IM \cite{liang2020we} & \ours \\
         \specialrule{1pt}{1pt}{1pt}
         \sc clipart & 25.04 & 30.88 \\
         \sc infograph & 21.58 & 26.44 \\
         \sc painting & 23.89 & 29.91 \\
         \sc quickdraw & 10.76 & 10.44 \\
         \sc real & 21.74 & 29.32 \\
         \sc sketch & 28.87 & 29.99 \\
         \bottomrule
         \sc Avg. & 21.98 & 26.13 \\
         \specialrule{1pt}{1pt}{1pt}
    \end{tabularx}
    \label{tab:domainnet}
\end{subfigure}
\end{table}

We compare our \ours with UDA and SFDA methods on multiple data sets for closed-set and open-set settings. First, we compare \ours with the baselines on the most common benchmark of {\sc office-home} for both closed-set and open-set settings. As can be seen from \cref{tab:cda-office-home} and \cref{tab:oda-office-home} we improve the performance over majority of the baselines. Especially, we consistently improve over SHOT-IM with our method. We also combine the nearest centroid pseudo-labelling, used in SHOT \cite{liang2020we}, with \ours (indicated as \ours\Plus\: in \cref{tab:cda-office-home} and \cref{tab:cda-visda}), and we find that it further helps improving the performance. Notably, the recently proposed $\text{A}^2\text{Net}$~\cite{xia2021adaptive} (which just addresses closed-set SFDA) outperforms our \ours in a couple of data sets, but uses a combination of several loss functions. Interplay of multiple losses can be hard to tune in practice. On the other hand, our method is simpler, more versatile and works for both the SFDA settings. Due to lack of space, we report the numbers for {\sc office31} in the appendix. Given the performance of the SFDA baseline methods in {\sc Office-home} and {\sc visda-c} are relatively high and closer to each other, the domain shift can be considered milder with respect to more challenging data set like {\sc domain-net}.

When we compare \ours with SHOT-IM on the challenging SFDA benchmark {\sc domain-net} the advantage of our \ours over SHOT-IM becomes imminent (\textit{cf.} \cref{tab:domainnet}), which is in line with the ablation study in~\cref{sec:abl}. Different from the previous data sets, the difficulty in mitigating domain-shift for {\sc domain-net} is evident from the low overall performance of both SHOT-IM and \ours. This data set can be seen as a real-world example of strong domain-shift. The improvement in the performance of \ours over SHOT-IM for {\sc domain-net} demonstrates that incorporating the uncertainty in the model's predictions plays a crucial role in SFDA. The conventional approach may overfit to noisy model predictions, leading to poor performance. Whereas, \ours can capture the uncertainty in predictions and down-weight the impact of noisy predictions.

\section{Discussion and Conclusion}
In this work, we demonstrated the need for uncertainty quantification in SFDA and proposed \ours for leveraging it during target adaptation. Our uncertainty-guided approach employs a Laplace approximation to the posterior, does not require specialized source training, and allows for efficient computation of predictive uncertainties. Our experiments showed that down-weighting \emph{distant} target data points in our novel uncertainty-weighted IM loss alleviates the misalignment of target data with the source hypothesis. We ran experiments on closed and open-set SFDA settings and show that \ours consistently improves upon the existing methods. Moreover, \ours has shown to be robust under mild distribution shifts and shows promising results under severe distribution shifts. 

While we mainly focused on the popular IM-based SFDA methods, our proposed uncertainty-guided adaptation is also applicable to other SFDA frameworks, \eg, neighbourhood clustering \cite{yang2021generalized} or extensions to the multi-source SFDA problem. Moreover, the principles we build upon are general, interpretable, and have strong backing in classical statistics. We believe that uncertainty-guided SFDA will become a backbone tool for future methods in DA that generalize over different problem domains, are less sensitive to the training setup, and will provide good results without extensive \emph{ad~hoc} tuning to each problem.

\smallskip

\noindent\textbf{Acknowledgements.} We acknowledge funding from EU H2020 projects SPRING (No.\ 871245) and AI4Media (No.\ 951911); the EUREGIO project OLIVER; Academy of Finland (No.\ 339730, 308640), and the Finnish Center for Artificial Intelligence (FCAI). We acknowledge the computational resources by the Aalto Science-IT project and CSC -- IT Center for Science, Finland. \sloppy

\clearpage

%
%
\bibliographystyle{splncs04}
\bibliography{egbib}

\clearpage
\appendix
{\centering\large\bf%
 Appendix: Uncertainty-guided Source-free Domain Adaptation%
}

\hfill

\setcounter{table}{0}
\renewcommand{\thetable}{A\arabic{table}}%
\setcounter{figure}{0}
\renewcommand{\thefigure}{A\arabic{figure}}%
\setcounter{equation}{0}
\renewcommand{\theequation}{A\arabic{equation}}%

This appendix is organised as follows:
Sec. \ref{app:notation} lists the notation used throughout the main text.
Sec. \ref{app:laplace} provides further details about Laplace approximations for approximate posterior inference in Bayesian neural networks.
Sec. \ref{app:algo} provides the algorithm of \ours.
Sec. \ref{sec:datasets} includes additional summary statistics on the data sets used for the empirical evaluation and lists additional results on the {\sc Office31} and {\sc visda-c} data set for the closed-set DA task.

\section{Notation}\label{app:notation}
The following notation is used throughout the paper:
\begin{table}[h]
    \centering
    \small
    \begin{tabularx}{0.68\textwidth}{ll}
    \toprule
    Notation & Description \\ \midrule
    $\data\tsource = \{(\vx_i\tsource,\vy_i\tsource)\}^{n\tsource}_{i=1}$ & Source data set \\
    $\data\ttarget = \{\vx_i\ttarget\}^{n\ttarget}_{i=1}$ & Target data set \\
    $\vx\tsource \in \X\tsource$ & Source inputs \\
    $\vy\tsource \in \Y\tsource$ & Source class labels \\
    $\vx\ttarget \in \X\ttarget$ & Target inputs \\
    $\labels\tsource, \labels\ttarget$ & Label sets \\
    $f, f'$ & Model functions (source and target) \\
    $g$ & Feature extractor\\
    $h$ & Hypothesis function\\
    $\beta, \theta$ & Parameterization of $f$ and $g$ \\
    $\vz = g(\vx)$ & Latent feature of observation $\vx$ \\
    $K$ & Number of classes \\
    $\phi_k(\cdot)$ & Softmax function \\
    $\MH$ & Hessian matrix \\
    \bottomrule
    \end{tabularx}
\end{table}

\section{Laplace Approximation} \label{app:laplace}
In Bayesian neural networks, we aim to incorporate uncertainty about the model and the model predictions. The standard approach places prior distributions ($p(\theta)$) onto the network parameters, which induces a probability distribution over the model predictions.
By conditioning the prior (in the weight-space) onto observed source data ($\data\tsource$), we obtain the posterior distribution over the network parameters $p(\theta \mid \data\tsource)$, allowing us to perform predictions by computing the posterior predictive distribution (see Eq. (4) in the main).

Let $\Psi(\theta)$ denote the unnormalised posterior distribution, \ie,
\begin{equation}
    \Psi(\theta) = p(\theta)\, p(\data\tsource \mid \theta) \, ,
\end{equation}
then the posterior distribution may be written as
\begin{equation} \label{eqn:app-posterior}
    p(\theta \mid \data\tsource) = \frac{1}{Z_\Psi} \Psi(\theta) \, ,
\end{equation}
where $Z_\Psi$ denotes the normalisation constant.
However, computing the posterior distribution and, subsequently, the posterior predictive distribution is intractable in general. We will, therefore, resort to a Laplace approximation to the posterior distribution.

Let $\theta_\MAP$ denote the maximum or a mode of the posterior distribution in \cref{eqn:app-posterior}. Then the second-order Taylor expansion of $\log \Psi(\theta)$ around $\theta_\MAP$ is given as:
\begin{equation}
\log \Psi(\theta) \approx \log \Psi(\theta_\MAP) - \frac{1}{2} \tanspos{(\theta - \theta_\MAP)}\, \MH\, (\theta - \theta_\MAP) \, ,
\end{equation}
where $\MH = -\nabla^2_\theta \log \Psi(\theta) \mid_{\theta=\theta_\MAP}$ is the negative Hessian of the log joint ($\log \Psi(\theta)$) evaluated at $\theta_\MAP$. Substituting the value of $\log \Psi(\theta)$ in \cref{eqn:app-posterior} gives us:
\begin{align}
& p(\theta \mid \data) = \frac{\Psi(\theta)}{\int \Psi(\theta) \dd \theta} \nonumber \\
&\approx \frac{ \Psi(\theta_\MAP) \exp \left(- \frac{1}{2} \tanspos{(\theta - \theta_\MAP)}\, \MH\, (\theta - \theta_\MAP)\right)}{ \Psi(\theta_\MAP) \int \exp \left(- \frac{1}{2} \tanspos{(\theta - \theta_\MAP)}\, \MH\, (\theta - \theta_\MAP)\right) \dd \theta} \nonumber \\
&= \frac{ \exp \left(- \frac{1}{2} \tanspos{(\theta - \theta_\MAP)}\, \MH\, (\theta - \theta_\MAP)\right)}{\int \exp \left(- \frac{1}{2} \tanspos{(\theta - \theta_\MAP)}\, \MH\, (\theta - \theta_\MAP) \right) \dd \theta} \,. \label{eqn:app-expanded-posterior}
\end{align}
The posterior can now be calculated in closed-form, and is given by:
\begin{align}
p(\theta \mid \data) &\approx \sqrt{\frac{\det \MH}{2\pi}} \exp \left( -\frac{1}{2} \tanspos{(\theta - \theta_\MAP)}\, \MH\, (\theta - \theta_\MAP) \right) \nonumber \\
&= \mathrm{N}(\theta \mid \mu_\MAP, \Sigma_\MAP), \label{eqn:app-lap}
\end{align}
where $\mu_\MAP = \theta_\MAP$ and $\Sigma_\MAP = \MH^{-1}$.

The posterior predictive distribution of an unseen datum $\vx\ttarget$ can now be approximated through Monte Carlo integration, \ie, 
\begin{equation}
    p(\vx\ttarget \mid \data\tsource) \approx \frac{1}{M} \sum^M_{j=1} p(\vx\ttarget \mid \theta_j) \, ,
\end{equation}
where $\theta_j \sim \mathrm{N}(\theta \mid \mu_\MAP, \Sigma_\MAP)$.

\section{Algorithm}\label{app:algo}

We report the pseudo-code for our \ours in Algo. \ref{algo:usfan}.

\begin{algorithm}[!h]
\setstretch{0.9}
\small
\SetAlgoLined\DontPrintSemicolon
\newcommand\mycommfont[1]{\textcolor{gray}{#1}}
\SetCommentSty{mycommfont}
\SetKwComment{Comment}{$\triangleright$\ }{}%
\SetKwComment{tcp}{$\triangleright$\ }{}%
\SetAlgoLined
\LinesNumbered
\SetKwInOut{in}{Input}
\SetKwInOut{out}{Output}
\in{A probabilistic source model $f = h \circ g$ with parameters $\{\beta\tsource_\MAP, \theta\tsource_\MAP, \MH^{-1}\}$, target data set $\data\ttarget$ containing $n\ttarget$ samples, mini-batch size $b$, temperature $\tau$, and $M$ MC steps.}

\out{Target-specific feature extractor parameters $\beta\ttarget$.}

\Repeat{converged}{
    
$\MX \gets$ sampleMiniBatch($\data\ttarget$, $b$) \;
$\MZ \gets g_{\beta\ttarget}(\MX)$\;
$\hat{\MY} \gets b \times K$ matrix of zeros\;

\tcp{Estimate predictive mean}
\For{$j = 1, \dots, M$}{
    $\theta_j \sim \mathrm{N}(\theta_j \mid \theta\tsource_\MAP, \MH^{-1}) $ \;
    $\hat{\MY} \gets \hat{\MY}$ + softmax($\nicefrac{h_{\theta_i}(\MZ)}{\tau}$)\;
}

$\hat{\MY} \gets \hat{\MY} / M$\;
\tcp{Compute model uncertainties}
\For{$i = 1, \dots, b$}{ 
    $w_i \gets \exp(-H(\hat{\vy_i}))$\;
}
Compute uncertainty-guided entropy \tcp*{Eq. (7)}
Compute divergence term \tcp*{Eq. (3)}
Compute \ours loss\;
Update parameters $\beta\ttarget$\;
} 

\caption{Uncertainty-guided Source-free DA}
\label{algo:usfan}
\end{algorithm}

\section{Data Set Details and Experiments} \label{sec:datasets}

We have summarized the statistics of the SFDA benchmark data sets used for the comparison against the state-of-the-art in \cref{tab:mtda_datasets}. To demonstrate the challenging aspect of having a strong domain-shift between the source and the target, we used the data set {\sc domain-net}. Moreover, the high number of semantic categories (345 classes) in {\sc domain-net} poses a challenge for the existing IM-based SFDA methods because of the lack of representative samples from every class in a given mini-batch.

\paragraph{Hyperparameter Selection.} We re-use the hyperparameters from the baseline of \cite{liang2020we}, \eg, the standard optimization technique for training such as SGD with an initial learning rate of $10^{-2}$ and $10^{-3}$ for ResNet-50 and ResNet-101, respectively. The learning rate is decayed by power decay \cite{ganin2015unsupervised}. We used the a batch size of 64 and we set $\alpha=0.1$ and $\gamma=0.5$. Exclusive to our method, we set the prior precision in LA equal to the weight decay, \ie $5 \cdot 10^{-4}$, and set the temperature $\tau=0.4$ for all our experiments.

\begin{table}[!h]
    \centering
    \small
    \tableoptions
    \caption{Data set summary for source-free domain adaptation}
    \begin{tabularx}{0.7\textwidth}{lccc}
         \toprule
         \sc Data set & \sc\#domains & \sc\#classes & \sc\#images \\
         \midrule
         \sc Office31 & 3 & 31 & 4,652 \\
         \sc Office-Home & 4 & 65 & 15,500 \\
         \sc Visda-C & 2 & 12 & $\sim$ 200K \\
         \sc Domain-Net & 6 & 345 & $\sim$ 0.6M \\
         \bottomrule
    \end{tabularx}
    \label{tab:mtda_datasets}
\end{table}

Additionally, we have reported the results of the experiments on {\sc office31} in \cref{tab:cda-office}. Similar to the results obtained on the other data sets reported in the main paper, \ours outperforms SHOT-IM on {\sc office31}. It must be noted that for data sets like {\sc office31}, the performance is already saturated, and the performance improvements of \ours over SHOT-IM are minor. Moreover, the data set shift is mild in most adaptation directions, evident from the saturated numbers. Thus, as discussed in the main paper, \ours does not yield remarkable improvement when the domain-shift is milder, and is most effective when much of the target data resides outside the source manifold. Nevertheless, when our method is combined with nearest centroid pseudo-labelling (like in SHOT), \ours{}+ further improve the performance. Through these extensive experiments on several SFDA benchmarks, we presented the advantages of our proposed method for the task of SFDA.

\begin{table}[h!]
    \centering
    \small
    \caption{Comparison of the classification accuracy on the {\sc Office31} for the closed-set SFDA using ResNet-50. Results on the small-scale {\sc office31} are known to be saturated. The visual appearance between the domains do not vary much, thus making the domain shift \textit{milder}. The improvement of \ours upon SHOT is moderate, but competitive w.r.t. $\text{A}^2\text{Net}$\cite{xia2021adaptive}, which requires complex training objectives}
    \begin{tabularx}{0.76\textwidth}{lcccccca} 
    \specialrule{1.5pt}{1pt}{1pt}
    \sc Method & A$\rightarrow$D & A$\rightarrow$W & D$\rightarrow$A & D$\rightarrow$W & W$\rightarrow$A & W$\rightarrow$D & \sc Avg. \\  
    \specialrule{1.5pt}{1pt}{1pt}
    ResNet-50 & 68.9 & 68.4 & 62.5 & 96.7 & 60.7 & 99.3 & 76.1 \\
    DANN \cite{ganin2016domain} & 79.7 & 82.0 & 68.2 & 96.9 & 67.4 & 99.1 & 82.2 \\
    DAN \cite{long2015learning} & 78.6 & 80.5 & 63.6 & 97.1 & 62.8 & 99.6 & 80.4 \\
    SAFN \cite{xu2019larger} & 90.7 & 90.1 & 73.0 & 98.6 & 70.2 & 99.8 & 87.1 \\
    CDAN \cite{long2018conditional} & 92.9 & 94.1 & 71.0 & 98.6 & 69.3 & 100. & 87.7 \\
    \hline
    SHOT-IM \cite{liang2020we} & 90.6 & 91.2 & 72.5 & 98.3 & 71.4 & 99.9 & 87.3 \\
    \ours(Ours) & 91.8 & 92.3 & 75.8 & 97.7 & 74.4 & 99.8 & 88.6\\
    $\text{A}^2\text{Net}$\cite{xia2021adaptive} & 94.5 & 94.0 & 76.7 & 99.2 & 76.1 & 100.0 & 90.1 \\
    \hline
    SHOT \cite{liang2020we} & 94.0 & 90.1 & 74.7 & 98.4 & 74.3 & 99.9 & 88.6 \\
    \ours\Plus\: (Ours) & 94.2 & 92.8 & 74.6 & 98.0 & 74.4 & 99.0 & 88.8 \\
    \specialrule{1.5pt}{1pt}{1pt}
    \end{tabularx}
    \label{tab:cda-office}
\end{table}

Due to lack of space in the main paper, in \cref{tab:cda-visda-full} we report the class-wise accuracy on the {\sc visda-c} data set, whose average accuracy has been reported in the Table 4 (a) of the main paper. While our \ours is competitive with SHOT-IM and SHOT, it underperforms with respect to $\text{A}^2\text{Net}$\cite{xia2021adaptive}. Nevertheless, \ours does not optimize a multitude of loss functions, making it more intuitive than the $\text{A}^2\text{Net}$.

\begin{table}[h!]
    \centering
    \small
    \tableoptions
    \setlength{\tabcolsep}{2.1pt}
    \caption{Comparison of the classification accuracy on the Visda-C for the closed-set DA, pertaining to the \textit{Synthetic $\rightarrow$ Real} direction, using ResNet-101. $\dagger$ indicates the numbers of \cite{liang2020we} that are obtained using the official code from the authors. Note that several SFDA methods perform equally well for {\sc visda-c}, hinting at saturating performance}
    \begin{tabularx}{\textwidth}{lcccccccccccca} 
    \specialrule{1.5pt}{1pt}{1pt}
    \sc Method & \rotatebox{90}{\sc plane} & \rotatebox{90}{\sc bcycl} & \rotatebox{90}{\sc bus} & \rotatebox{90}{\sc car} & \rotatebox{90}{\sc horse} & \rotatebox{90}{\sc knife} & \rotatebox{90}{\sc mcycl} & \rotatebox{90}{\sc person} & \rotatebox{90}{\sc plant} & \rotatebox{90}{\sc sktbrd} & \rotatebox{90}{\sc train} & \rotatebox{90}{\sc truck} & \rotatebox{90}{\sc Avg.} \\  
    \specialrule{1.5pt}{1pt}{1pt}
    ResNet-101 & 55.1 & 53.3 & 61.9 & 59.1 & 80.6 & 17.9 & 79.7 & 31.2 & 81.0 & 26.5 & 73.5 & 8.5 & 52.4 \\
    DANN \cite{ganin2016domain} &  81.9 & 77.7 & 82.8 & 44.3 & 81.2 & 29.5 & 65.1 & 28.6 & 51.9 & 54.6 & 82.8 & 7.8 & 57.4 \\
    ADR \cite{saito2017adversarial} & 94.2 & 48.5 & 84.0 & 72.9 & 90.1 & 74.2 & 92.6 & 72.5 & 80.8 & 61.8 & 82.2 & 28.8 & 73.5 \\
    CDAN \cite{long2018conditional} & 85.2 & 66.9 & 83.0 & 50.8 & 84.2 & 74.9 & 88.1 & 74.5 & 83.4 & 76.0 & 81.9 & 38.0 & 73.9 \\
    CDAN+BSP \cite{chen2019transferability} & 92.4 & 61.0 & 81.0 & 57.5 & 89.0 & 80.6 & 90.1 & 77.0 & 84.2 & 77.9 & 82.1 & 38.4 & 75.9 \\
    SAFN \cite{xu2019larger} & 93.6 & 61.3 & 84.1 & 70.6 & 94.1 & 79.0 & 91.8 & 79.6 & 89.9 & 55.6 & 89.0 & 24.4 & 76.1 \\
    SWD \cite{lee2019sliced} & 90.8 & 82.5 & 81.7 & 70.5 & 91.7 & 69.5 & 86.3 & 77.5 & 87.4 & 63.6 & 85.6 & 29.2 & 76.4 \\
    \hline
    DANCE \cite{saito2020universal} & - & - & - & - & - & - & - & - & - & - & - & - & 70.2 \\
    $\text{SHOT-IM}^{\dagger}$ \cite{liang2020we} & 94.2 & 87.6 & 78.6 & 48.6 & 92.1 & 92.9 & 76.4 & 76.2 & 89.4 & 86.6 & 88.8 & 52.7 & 80.3 \\
    \ours (Ours) & 95.1 & 87.0 & 76.8 & 50.1 & 92.9 & 94.3 & 79.0 & 78.0 & 88.4 & 87.5 & 87.7 & 57.3 & 81.2 \\
    3C-GAN \cite{li2020model} & 94.8 & 73.4 & 68.8 & 74.8 & 93.1 & 95.4 & 88.6 & 84.7 & 89.1 & 84.7 & 83.5 & 48.1 & 81.6 \\
    $\text{A}^2\text{Net}$\cite{xia2021adaptive} & 94.0 & 87.8 & 85.6 & 66.8 & 93.7 & 95.1 & 85.8 & 81.2 & 91.6 & 88.2 & 86.5 & 56.0 & 84.3 \\
    \hline
    $\text{SHOT}^{\dagger}$ \cite{liang2020we} & 94.9 & 87.1 & 76.9 & 55.0 & 94.2 & 95.4 & 80.8 & 80.0 & 89.5 & 88.7 & 85.6 & 60.5 & 82.4 \\
    \ours + (Ours) & 94.9 & 87.4 & 78.0 & 56.4 & 93.8 & 95.1 & 80.5 & 79.9 & 90.1 & 90.1 & 85.3 & 60.4 & 82.7 \\
    \specialrule{1.5pt}{1pt}{1pt}
    \end{tabularx}
    \label{tab:cda-visda-full}
\end{table}

\end{document}